\documentclass[conference]{IEEEtran}
\IEEEoverridecommandlockouts
\usepackage{cite}
\usepackage{amsmath,amssymb,amsfonts}
\usepackage{algorithmic}
\usepackage{graphicx}
\usepackage{textcomp}
\usepackage{xcolor}
\usepackage{comment}
\usepackage{tikz}
\usepackage[absolute,overlay]{textpos}
\usepackage{blindtext}
\usepackage{relsize}
\usepackage{hyperref}

\tikzset{fontscale/.style={font=\relsize{#1}}}

\def\BibTeX{{\rm B\kern-.05em{\sc i\kern-.025em b}\kern-.08em
    T\kern-.1667em\lower.7ex\hbox{E}\kern-.125emX}}

\begin{document}
\title{HSI-Drive v2.0: More Data for New Challenges in Scene Understanding for Autonomous Driving*\\
\thanks{*This work was partially supported by the Basque Government under grants PRE\_2022\_2\_0210 and KK-2023/00090, by the Spanish Ministry of Science and Innovation under grant PID2020-115375RB-I00 and by the University of the Basque Country (UPV-EHU) under grant GIU21/007.}
}

\author{\IEEEauthorblockN{1\textsuperscript{st} Jon Gutiérrez-Zaballa}
\IEEEauthorblockA{\textit{Department of Electronics Technology} \\
\textit{University of the Basque Country}\\
Bilbao, Spain \\
0000-0002-6633-4148}
\and
\IEEEauthorblockN{2\textsuperscript{nd} Koldo Basterretxea}
\IEEEauthorblockA{\textit{Department of Electronics Technology} \\
\textit{University of the Basque Country}\\
Bilbao, Spain \\
0000-0002-5934-4735}
\and
\IEEEauthorblockN{3\textsuperscript{rd} Javier Echanobe}
\IEEEauthorblockA{\textit{Department of Electricity and Electronics} \\
\textit{University of the Basque Country}\\
Leioa, Spain \\
0000-0002-1064-2555}
\and
\IEEEauthorblockN{4\textsuperscript{th} M. Victoria Martínez}
\IEEEauthorblockA{\textit{Department of Electricity and Electronics} \\
\textit{University of the Basque Country}\\
Leioa, Spain}
\and
\IEEEauthorblockN{5\textsuperscript{rd} Unai Martinez-Corral}
\IEEEauthorblockA{\textit{Department of Electronics Technology} \\
\textit{University of the Basque Country}\\
Bilbao, Spain \\
0000-0003-1752-9181}
}

\maketitle
\begin{textblock*}{21cm}(1.5cm,26cm)
  \begin{tikzpicture}
    \draw (0,0) rectangle (18,0.5); 
    \end{tikzpicture}
\end{textblock*} 

\begin{textblock*}{21cm}(0cm,26cm)
  \begin{tikzpicture}
    \node (center) {c};
    \path (center)+(10.5,4) node [fontscale=-1] (name) {\copyright 2024 IEEE. Final published version of the article can be found at \href{https://ieeexplore.ieee.org/document/10371793}{10.1109/SSCI52147.2023.10371793}.};
    \end{tikzpicture}
\end{textblock*}

\begin{abstract}
We present the updated version of the HSI-Drive dataset aimed at developing automated driving systems (ADS) using hyperspectral imaging (HSI).
The v2.0 version includes new annotated images from videos recorded during winter and fall in real driving scenarios.
Added to the spring and summer images included in the previous v1.1 version, the new dataset contains 752 images covering the four seasons.
In this paper, we show the improvements achieved over previously published results obtained on the v1.1 dataset, showcasing the enhanced performance of models trained on the new v2.0 dataset.
We also show the progress made in comprehensive scene understanding by experimenting with more capable image segmentation models.
These models include new segmentation categories aimed at the identification of essential road safety objects such as the presence of vehicles and road signs, as well as highly vulnerable groups like pedestrians and cyclists.
In addition, we provide evidence of the performance and robustness of the models when applied to segmenting HSI video sequences captured in various environments and conditions.
Finally, for a correct assessment of the results described in this work, the constraints imposed by the processing platforms that can sensibly be deployed in vehicles for ADS must be taken into account.
Thus, and although implementation details are out of the scope of this paper, we focus our research on the development of computationally efficient, lightweight ML models that can eventually operate at high throughput rates.
The dataset and some examples of segmented videos are available in https://ipaccess.ehu.eus/HSI-Drive/.
\end{abstract}


\begin{table*}[!t]
\centering
\caption{Frequency of each of the classes in the HSI Drive v2.0 dataset.}
\resizebox{17.5cm}{!}{%
\begin{tabular}{cccccccccccc}
  \cline{2-12}
  \multicolumn{1}{c|}{} &
  \multicolumn{1}{c|}{\textbf{Total}} &
  \multicolumn{1}{c|}{\textbf{Road}} &
  \multicolumn{1}{c|}{\textbf{R. Marks $^{\mathrm{a}}$}} &
  \multicolumn{1}{c|}{\textbf{Veg.$^{\mathrm{b}}$}} &
  \multicolumn{1}{c|}{\textbf{Pain. Met. $^{\mathrm{c}}$}} &
  \multicolumn{1}{c|}{\textbf{Sky}} &
  \multicolumn{1}{c|}{\textbf{Concrete}} &
  \multicolumn{1}{c|}{\textbf{Ped. $^{\mathrm{d}}$}} &
  \multicolumn{1}{c|}{\textbf{Water}} &
  \multicolumn{1}{c|}{\textbf{Unpain. Met. $^{\mathrm{e}}$}} &
  \multicolumn{1}{c|}{\textbf{Glass}} \\ \hline
  \multicolumn{1}{|c|}{\textbf{Num. pixels}} &
  \multicolumn{1}{c|}{43 947 503} &
  \multicolumn{1}{c|}{26 690 619} &
  \multicolumn{1}{c|}{1 325 343} &
  \multicolumn{1}{c|}{9 339 224} &
  \multicolumn{1}{c|}{948 852} &
  \multicolumn{1}{c|}{2 511 496} &
  \multicolumn{1}{c|}{2 315 153} &
  \multicolumn{1}{c|}{209 531} &
  \multicolumn{1}{c|}{12 330} &
  \multicolumn{1}{c|}{348 341} &
  \multicolumn{1}{c|}{246 614} \\ \hline
  \multicolumn{1}{|c|}{\textbf{\%}} &
  \multicolumn{1}{c|}{100} &
  \multicolumn{1}{c|}{60.73} &
  \multicolumn{1}{c|}{3.02} &
  \multicolumn{1}{c|}{21.25} &
  \multicolumn{1}{c|}{2.16} &
  \multicolumn{1}{c|}{5.71} &
  \multicolumn{1}{c|}{5.27} &
  \multicolumn{1}{c|}{0.48} &
  \multicolumn{1}{c|}{0.03} &
  \multicolumn{1}{c|}{0.79} &
  \multicolumn{1}{c|}{0.56} \\ \hline
  \multicolumn{12}{l}{$^{\mathrm{a}}$Road Marks. $^{\mathrm{b}}$ Vegetation. $^{\mathrm{c}}$ Painted Metal. $^{\mathrm{d}}$ Pedestrian. $^{\mathrm{e}}$ Unpainted Metal.}
\end{tabular}}
\label{tab:datasetPartition}
\end{table*}

\section{Introduction}
The exploration of hyperspectral imaging (HSI) processing techniques in the development of autonomous driving systems (ADS) and advanced driver assistance systems (ADAS) is now possible due to the availability of small-size, snapshot hyperspectral cameras that enable the recording of hyperspectral images at video rates from moving platforms \cite{basterretxeahsi, mv1-d2048x1088-hs02-96-g2}.
However, there are inherent technological constraints and engineering challenges associated with acquiring and processing spectral data at video rates in real driving conditions since outdoor recording implies dealing with varying lighting and weather conditions, the presence of fast moving objects, etc.
Processing the spectral information contained in such images implies handling a variety of non-controlled natural illumination and backgrounds, sensor saturation effects, the presence of objects at very different distances and sometimes severe spectral mixing due to sensor technology and limited spatial resolution.
To address these challenges in intelligent vision applications, spectral data need to be preprocessed and complemented with relevant spatial information.

Deep learning models, particularly fully convolutional networks (FCNs), have demonstrated outstanding performance in capturing spatial features of objects with various sizes and shapes and have been widely applied to the segmentation of hyperspectral images \cite{phan2022efficient, taghizadeh2011comparison, seidlitz2022robust, fricker2019convolutional}.
The availability of large volumes of data is crucial for the development of robust deep learning models trained on datasets with high data variability.
Unfortunately, there are only a few datasets specifically designed to train and test HSI-processing ML systems for the development of ADS \cite{basterretxeahsi,winkens2017hyko, hsiroad, you2019hyperspectral}.
In particular, HSI-Drive \cite{ipaccess} is a structured HSI dataset that is being used for the research of hyperspectral image segmentation systems to be deployed as ADAS in automobiles.
In this paper, we present the extended version of HSI-Drive database (v2.0), which contains more than double the data than the previous v1.1 version.
We show how the availability of more data acquired in more diverse environments allows to develop more accurate and robust HSI segmentation models, as well as to widen the capabilities of the HSI processing systems for a more comprehensive scene understanding.

The remainder of the paper is organized as follows: Section \ref{sec:dataset} provides detailed information about the updates in the new version of the HSI-Drive dataset.
Section \ref{sec:experimentalsetup} presents the experimental setup, including data partitioning, preprocessing, FCN model development, and the rationale behind each experiment.
Section \ref{sec:segmentationresults} presents the segmentation metrics for the different experiments.
Additionally, it includes illustrations of the segmentation system performance, showcasing the evaluation of representative driving scenes.
Finally, Section \ref{sec:conclu} concludes the paper and discusses potential future work.

\section{HSI-Drive v2.0}\label{sec:dataset}
The v2.0 version of the HSI-Drive dataset \cite{ipaccess}, released in December 2022, contains 752 manually labeled images from recordings made in fall (201 images), winter (206 images), spring (166 images) and summer (155 images).
Compared to the previous v1.1 dataset, which contains 276 images recorded only in summer and spring, v2.0 provides an increase of more than 272{\textbf{\%}} in the total number of images and a great improvement in terms of data diversity.
The dataset contains almost 44 million labeled pixels, categorized into 10 classes as shown in Table \ref{tab:datasetPartition}.
Despite the labeling being primarily aimed to benefit spectral classification, categories have been defined to be significant to the scope of application, hence most of them comprise different materials.
In consequence, each class exhibits very different spectral variability, which challenges inter-class separability.
For instance, while the Road category encompasses only tarmac surfaces, the Pedestrian category includes individuals such as passers-by, cyclists, motorcyclists and animals.
On the other hand, the careful structuring of the dataset according to season of the year, weather conditions, daytime and road type provides two potential avenues for research: developing general and robust classification systems that remain unaffected by the diversity of lighting and environmental conditions, and selecting a specific subset of the dataset to study phenomena closely associated with particular driving and environmental situations.

The images in the dataset were captured using a Photonfocus camera equipped with an Imec 25-band VIS-NIR (535nm-975nm) mosaic spectral filter on a CMOSIS CMV200 image wafer sensor \cite{mv1-d2048x1088-hs02-96-g2}.
The raw images in the dataset have a spatial resolution of 1088 x 2048 pixels, with each pixel measuring 5$\mu m$ x 5$\mu m$.
However, the spectral bands are extracted from a mosaic formed by 5x5 pixel window Fabri-Perot filters, resulting in a reduced resolution output cube with 216 x 409 x 25 size.
The images were recorded with a digital resolution of 12 bits, leading to an estimated signal-to-noise ratio (SNR) ranging between 23.43dB and 27.29dB for the recording setups used.

Acquiring images from a moving vehicle under varying lighting conditions presents several challenges.
First, to avoid motion blur, an appropriate exposure-time limit has to be set.
This limit, in turn, challenges the acquisition of images under low lighting conditions.
Adjusting the sensor's gain can partially compensate for the lack of light, but it also amplifies the noise in the image data.
The f-number (aperture) of the camera optics can also be readjusted to increase the reception of light, but this affects the depth of field and the angle of incidence of the light beams which, at the same time, produces variations in the response of the Fabry-Perot filters of the sensor.
Secondly, in sunny conditions with significant light contrasts between illuminated and shadowed surfaces, setting the exposure-time becomes crucial to minimize or prevent pixel saturation, which occurs due to the sensor's limited dynamic range.
In the end, increasing the number of different camera configurations results in a more burdensome and time-consuming image preprocessing pipeline in order to preserve the coherence of the spectral information of images which, at the same time, may compromise the compliance with real-time operation requirements of ADS/ADAS.

\section{Experimental setup}\label{sec:experimentalsetup}
\subsection{Segmentation experiments}
In this section, we present four experiments on HSI-based semantic segmentation using HSI-Drive 2.0 data.
Two experiments (3- and 5-classes) have been previously conducted in earlier studies \cite{gutierrez22, gutierrez2023chip} and FCN models have been updated and improved using the new data.
The two new experiments involve 6-class segmentation and expand upon the 5-class experiment by including the categories Painted Metal and Pedestrian respectively.
The purpose of these additions is to enhance the overall understanding of the environment perceived by the system, thereby contributing to improve scene comprehension.
As described below, obtained experimental results demonstrate that incorporating new training data enhances the classification capabilities, performance and robustness of the developed segmentation systems.

Experiment 1 was designed to perform a simple segmentation of the Road (tarmac) and the Road Marks in the scenes.
This set-up is particularly useful for lane-keeping and trajectory planning systems.
In Experiment 2, additional information about the background is incorporated by including Sky and Vegetation categories.
This extension enables the identification of potential obstacles such as vehicles, cyclists, pedestrians, etc., which may demand responsive actions.
Furthermore, the segmentation reveals the presence of road signs, traffic lights and information panels located at the sides and above the roads.

The newly designed Experiment 3 incorporates the segmentation of Painted Metal surfaces.
This category specifically focuses on the presence of vehicles and traffic signs, which could help to improve systems for signal identification, emergency braking, collision alerts, and adaptive cruise control.
Experiment 4 aims to cover the segmentation of pedestrians, cyclists and motorcyclists, whose effective identification is the prerequisite for their protection in ADS.

\subsection{Data partition and preprocessing}
The 752 images were divided into 5 subsets for a 5-fold cross-validation training scheme.
The partitioning was performed based on a proportionality criterion considering the distribution of the images across the dataset structure, i.e. daytime, climatology, season and road type.
To prevent local overfitting and improve the generalization performance of the models, a validation subset was used for early stopping in training.
Specifically, 3 subsets were used for training (60\%), 1 for validation (20\%), and 1 for testing (20\%).
To mitigate the influence of random weight initialization, each training was repeated 3 times.

Regarding raw image preprocessing stage, we performed image cropping, reflectance correction through dark and flat images, and partial demosaicing by spatial bilinear interpolation (see \cite{gutierrez2023chip} for further details).
We have removed the median filtering step included in previous experiments since it was observed that spatial filtering does not yield any improvements when training models that incorporate convolutional spatial filters.
Finally, to enhance image invariance to lighting conditions (shadow removal), a per-pixel normalization (dividing each pixel's value by the sum of its spectral signature) is performed at the end of the preprocessing pipeline, as described in \cite{gevers2000colour}, which extends the work from \cite{finlayson2005removal} to the hyperspectral domain.

\subsection{Model training and optimization}
In this work, we continue to explore encoder-decoder FCN models to effectively combine spectral and spatial features for the semantic segmentation of HSI.
Compared to the tiny FCN models reported in \cite{gutierrez2023chip}, we have explored deeper encoder structures to make the most of the availability of new data and perform the segmentation of the whole images in a single pass.
Training on larger images implies using deeper networks to effectively extract spatial features at different scales.

The models were trained on a NVIDIA GFORCE RTX-3090 with 24GB of memory.
During training, a batch size of 23 images was utilized, while a batch size of 49 images was used for validation.
Best fitting was obtained for an Adam optimizer with an initial learning rate of 0.001, gradient decay factor of 0.9, squared gradient decay factor of 0.999, 200 epochs, and data shuffling at each epoch.
The objective function was an inverse-frequency weighted cross-entropy loss to ensure higher weights for the minority classes.

A grid search hyperparameter optimization study was conducted to search for the best trade-off between model complexity and classification performance.
Explored model hyperparameters were the encoder-depth (2, 3, 4, and 5), the input image-size (whole image versus image patching), the number of filters in the input convolutional layer (8, 16, and 32), the size of convolutional kernels (3 and 5), and the dropout layer placement (after each encoder block or only after the first and last ones) and dropout rates (0, 0.2, 0.5).
During training, regularization techniques were applied to the convolutional filters and three different learning rates (0.01, 0.001, 0.0001) were also essayed.
The resulting optimum model, which is a modification of the architecture shown in Fig. 6 of \cite{gutierrez2023chip},  is composed of 32 filters in the first convolutional block, an encoder depth of 5 layers, and 3x3 convolutional kernels.
Since a stride value of 2 in the pooling layers constraints the input image size to be a multiple of 2 raised to the encoder depth, the largest compatible size is 192x384 so, during training, each 216x409 image is divided into four 192x384 overlapping patches.
During testing, patches can be merged to recover the original size if necessary.

The model contains a total of 31.10 million parameters and requires 34.87 giga floating-point operations (GFLOPS) per inference.
In order to meet the demanding latency and memory footprint implementation constraints of ADAS/ADS systems, we simplified the model by applying an iterative pruning algorithm based on the analysis of the computational complexity of each layer and the evaluation of the model's accuracy.
As a result of this optimization process, the computational load was reduced to 8.49 GFLOPS and the number of parameters to only 320K with no noticeable impact on the model's accuracy, even after 8-bit integer quantization was performed.
The detailed description of the procedure followed to achieve this remarkable model compression is out of the scope of this paper and will be published in a near future.

\section{Results}\label{sec:segmentationresults}
Tables \ref{tab:exp1} to \ref{tab:exp5} show the segmentation metrics (Recall, Precision, and Intersection over Union, IoU) for complete 216x409 images in each experiment.
Global metrics consider the frequency of each class in the dataset, while weighted metrics consider the inverse frequency of each class in the dataset, prioritizing minority classes.
The formulas used to calculate the metrics can be found in \cite{gutierrez22}.

\subsection{Segmentation results}
In experiment 1, class division was: Road - 60.73\%, Road Marks - 3.02\% and No Drivable - 36.25\%.
The results presented in Table \ref{tab:exp1} depict significant improvements compared to the previous models trained on the v1.1 dataset.
The overall IoU shows a notable increase from 91.50 to 96.87, while the weighted IoU improves from 72.60 to 88.55.
Particularly, the precision of the Road Marks class substantially increases from 77.22 to 95.53.
Moreover, as we will discuss later, the satisfactory performance of the network extends robustly to unlabeled pixels, as observed in video sequences.

\begin{table}[t]
\centering
\caption{Segmentation metrics for Exp. 1: three classes}
\label{tab:exp1}
\resizebox{7.5cm}{!}{%
\begin{tabular}{c|c|c|c|}
\cline{2-4}
& \multicolumn{3}{|c|}{\textbf{Mean} $\pm$ \textbf{Std}} \\ \hline
\multicolumn{1}{|c|}{\textbf{Metric}} & \textbf{Recall} & \textbf{Precision} & \textbf{IoU} \\ \hline
\multicolumn{1}{|c|}{\textbf{Road}}       & 99.20 $\pm$ 0.48 & 98.34 $\pm$ 0.61 & 97.57 $\pm 0.38 $ \\ \hline
\multicolumn{1}{|c|}{\textbf{Road Marks}} & 91.09 $\pm$ 2.53 & 95.53 $\pm$ 1.04 & 87.39 $\pm 2.94 $ \\ \hline
\multicolumn{1}{|c|}{\textbf{No Dri}}     & 97.67 $\pm$ 1.02 & 98.75 $\pm$ 0.89 & 96.47 $\pm 0.49 $ \\ \hline
\multicolumn{1}{|c|}{\textbf{Global}}     & 98.40 $\pm$ 0.24 & 98.40 $\pm$ 0.24 & 96.87 $\pm 0.47 $ \\ \hline
\multicolumn{1}{|c|}{\textbf{Weighted}}   & 91.98 $\pm$ 2.21 & 95.92 $\pm$ 0.92 & 88.55 $\pm 2.60 $ \\ \hline
\end{tabular}}
\end{table}

In experiment 2, class division was: Road - 60.73\%, Road Marks - 3.02\%, Vegetation - 21.25\%, Sky - 5.71\% and Other - 9.29\%.
As shown in Table \ref{tab:exp2}, the addition of two new classes with good separability indexes (Vegetation and Sky) does not penalize the accuracy for other minority classes.
Again, there is a significant improvement when compared to the results obtained on the v1.1 dataset, with the global IoU increasing from 87.66 to 94.51, and the weighted IoU rising from 75.93 to 87.18.
This improvement is mainly attributed to the increase in IoU of the Road Marks class from 64.90 to 86.08.

\begin{table}[b]
\centering
\caption{Segmentation metrics for Exp. 2: five classes}
\label{tab:exp2}
\resizebox{7.5cm}{!}{%
\begin{tabular}{c|c|c|c|}
\cline{2-4}
& \multicolumn{3}{|c|}{\textbf{Mean} $\pm$ \textbf{Std}} \\ \hline
\multicolumn{1}{|c|}{\textbf{Metric}} & \textbf{Recall} & \textbf{Precision} & \textbf{IoU} \\ \hline
\multicolumn{1}{|c|}{\textbf{Road}}       & 99.35 $\pm$ 0.33 & 98.11 $\pm$ 0.55 & 97.49 $\pm$ 0.47 \\ \hline
\multicolumn{1}{|c|}{\textbf{Road Marks}} & 90.85 $\pm$ 1.48 & 94.23 $\pm$ 2.34 & 86.08 $\pm$ 2.93 \\ \hline
\multicolumn{1}{|c|}{\textbf{Vegetation}} & 97.84 $\pm$ 0.67 & 96.88 $\pm$ 1.24 & 94.86 $\pm$ 1.53 \\ \hline
\multicolumn{1}{|c|}{\textbf{Sky}}        & 92.49 $\pm$ 2.46 & 98.55 $\pm$ 0.40 & 91.24 $\pm$ 2.20 \\ \hline
\multicolumn{1}{|c|}{\textbf{Other}}      & 85.75 $\pm$ 3.36 & 91.01 $\pm$ 2.71 & 78.96 $\pm$ 1.79 \\ \hline
\multicolumn{1}{|c|}{\textbf{Global}}     & 97.12 $\pm$ 0.41 & 97.10 $\pm$ 0.42 & 94.51 $\pm$ 0.75 \\ \hline
\multicolumn{1}{|c|}{\textbf{Weighted}}   & 91.18 $\pm$ 1.29 & 95.10 $\pm$ 1.41 & 87.18 $\pm$ 2.02 \\ \hline
\end{tabular}}
\end{table}

In experiment 3, class division was: Road - 60.73\%, Road Marks - 3.02\%, Vegetation - 21.25\%, Painted Metal - 2.16\%, Sky - 5.71\% and Other - 7.13\%.
As shown in Table \ref{tab:exp3}, while the mean precision of the Painted Metal class is 85.20\%, the recall value of 65.40\% requires improvement.
Nevertheless, due to its heterogeneous nature and high intra-class variability, a more detailed analysis of the segmented images is required in order to determine whether classification successes and failures occur consistently or under particular lighting or weather conditions.

\begin{table}[!b]
\centering
\caption{Segmentation metrics for Exp 3: six classes.}
\label{tab:exp3}
\resizebox{7.5cm}{!}{%
\begin{tabular}{c|c|c|c|}
\cline{2-4}
 & \multicolumn{3}{|c|}{\textbf{Mean $\pm$ Std}} \\ \hline
\multicolumn{1}{|c|}{\textbf{Metric}} & \textbf{Recall} & \textbf{Precision} & \textbf{IoU} \\ \hline
\multicolumn{1}{|c|}{\textbf{Road}}             & 99.19 $\pm$ 0.55 & 98.13 $\pm$ 0.51 & 97.34 $\pm$ 0.36 \\ \hline
\multicolumn{1}{|c|}{\textbf{Road Marks}}       & 91.42 $\pm$ 1.52 & 92.56 $\pm$ 1.90 & 85.20 $\pm$ 2.43 \\ \hline
\multicolumn{1}{|c|}{\textbf{Vegetation}}       & 98.26 $\pm$ 0.92 & 95.45 $\pm$ 2.04 & 93.84 $\pm$ 1.81 \\ \hline
\multicolumn{1}{|c|}{\textbf{Painted Metal}}    & 65.40 $\pm$ 4.84 & 85.20 $\pm$ 5.70 & 58.61 $\pm$ 4.44 \\ \hline
\multicolumn{1}{|c|}{\textbf{Sky}}              & 95.24 $\pm$ 2.34 & 96.96 $\pm$ 1.99 & 92.30 $\pm$ 1.11 \\ \hline
\multicolumn{1}{|c|}{\textbf{Other}}            & 77.85 $\pm$ 5.76 & 85.48 $\pm$ 1.71 & 68.65 $\pm$ 4.17 \\ \hline
\multicolumn{1}{|c|}{\textbf{Global}}           & 95.48 $\pm$ 2.25 & 96.17 $\pm$ 0.61 & 93.07 $\pm$ 1.03 \\ \hline
\multicolumn{1}{|c|}{\textbf{Weighted}}         & 79.26 $\pm$ 4.59 & 89.59 $\pm$ 2.95 & 74.45 $\pm$ 3.05 \\ \hline
\end{tabular}}
\end{table}

In experiment 4, class division was: Road - 60.73\%, Road Marks - 3.02\%, Vegetation - 21.25\%, Pedestrian - 0.48\%, Sky - 5.71\% and Other - 8.81\%.
The precision obtained for the Pedestrian class is similar to that obtained in Experiment 3 for the Painted Metal class but with a better recall (mean 70.20).

\begin{table}[!b]
\centering
\caption{Segmentation metrics for Exp 4: six classes.}
\label{tab:exp4}
\resizebox{7.5cm}{!}{%
\begin{tabular}{c|c|c|c|}
\cline{2-4}
 & \multicolumn{3}{|c|}{\textbf{Mean $\pm$ Std}} \\ \hline
\multicolumn{1}{|c|}{\textbf{Metric}} & \textbf{Recall} & \textbf{Precision} & \textbf{IoU} \\ \hline
\multicolumn{1}{|c|}{\textbf{Road}}             & 99.02 $\pm$ 0.31 & 98.00 $\pm$ 0.79 & 97.04 $\pm$ 0.80 \\ \hline
\multicolumn{1}{|c|}{\textbf{Road Marks}}       & 87.87 $\pm$ 4.05 & 91.66 $\pm$ 2.00 & 81.85 $\pm$ 4.37 \\ \hline
\multicolumn{1}{|c|}{\textbf{Vegetation}}       & 98.34 $\pm$ 0.54 & 95.07 $\pm$ 2.22 & 93.56 $\pm$ 2.02 \\ \hline
\multicolumn{1}{|c|}{\textbf{Pedestrian}}       & 70.02 $\pm$ 3.64 & 84.26 $\pm$ 8.15 & 61.94 $\pm$ 1.41 \\ \hline
\multicolumn{1}{|c|}{\textbf{Sky}}              & 91.86 $\pm$ 9.78 & 97.44 $\pm$ 1.01 & 89.23 $\pm$ 8.66 \\ \hline
\multicolumn{1}{|c|}{\textbf{Other}}            & 80.59 $\pm$ 6.87 & 89.83 $\pm$ 2.84 & 74.20 $\pm$ 7.07 \\ \hline
\multicolumn{1}{|c|}{\textbf{Global}}           & 96.42 $\pm$ 1.07 & 96.37 $\pm$ 1.10 & 93.26 $\pm$ 1.97 \\ \hline
\multicolumn{1}{|c|}{\textbf{Weighted}}         & 74.45 $\pm$ 3.48 & 86.49 $\pm$ 6.35 & 67.13 $\pm$ 1.47 \\ \hline
\end{tabular}}
\end{table}

\subsection{Influence of lighting and weather conditions}
The structured organization of the HSI-Drive dataset allows for defining subsets of data that were obtained under similar circumstances.
Here we present the results of an experiment aimed to analyze the consequences of training the FCN with such subsets and t explore the performance of the FCN under different conditions.
This analysis provides insights into the conditions that may be more demanding for the segmentation system and serve as a valuable guide for future research efforts to address those specific conditions.

\begin{table}[b]
\centering
\caption{Performance comparison for different lighting and weather conditions in Exp. 2.}
\label{tab:exp5}
\resizebox{8cm}{!}{
\begin{tabular}{|c|c|c|c|c|c|c|}
\hline
\multicolumn{1}{|c|}{\textbf{Condition}} & \multicolumn{1}{c|}{\textbf{Dawn}} & \multicolumn{1}{c|} {\textbf{Midday}} & \multicolumn{1}{c|}{\textbf{Sunset}} & \multicolumn{1}{c|}{\textbf{Sunny}} & \multicolumn{1}{c|}{\textbf{Cloudy}} & \multicolumn{1}{c|}{\textbf{Rainy}} \\ \hline
\textbf{Road}       & 97.16 & 96.86 & 97.27 & 95.47 & 98.19 & 97.11 \\ \hline
\textbf{Road Marks} & 78.93 & 85.03 & 84.87 & 76.69 & 91.48 & 75.99 \\ \hline
\textbf{Vegetation} & 94.33 & 97.74 & 94.37 & 92.72 & 97.82 & 97.51 \\ \hline
\textbf{Sky}        & 91.22 & 87.68 & 90.88 & 89.58 & 96.47 & 92.95 \\ \hline
\textbf{Others}     & 80.45 & 80.33 & 72.89 & 73.76 & 80.49 & 78.02 \\ \hline
\textbf{Global}     & 94.22	& 94.46 & 94.08 & 91.84 & 96.29 & 94.78 \\ \hline
\textbf{Weighted}   & 84.06 & 86.46 & 84.67 & 80.35 & 91.61 & 83.37 \\ \hline
\end{tabular}}
\end{table}

According to the results shown in Table \ref{tab:exp5}, some conclusions can be drawn.
Regarding  weather conditions, the FCN achieves the best performance for the Cloudy subset.
This is consistent with the more favorable and homogeneous illumination conditions, with lighter shadows and less overexposure.
Rainy conditions are supposed to be more challenging due to the reduced visibility and the high probability of the presence of glares and light reflections, and condensation and water drops on the lens.
Unexpectedly, there is no significant reduction in the general performance metrics compared to other conditions, except for the Road Marks.
The poorest results were obtained in the Sunny subset, which contains images with severe illumination contrast that combine very low reflectance values in the shadows with overexposed areas in the sunny areas.

Regarding lighting condition variability throughout the day, the Midday subset, characterized by sufficient lighting regardless of weather conditions and a more zenithal position of the sun, yields the best results.
In contrast, both the Dawn and Sunset subsets are the most challenging ones, as they often contain images with severe glares, high contrasts, and low lighting.
However, there are no noticeable differences in the global and weighted index values.

\subsection{Detailed evaluation of some representative scenes}
Although evaluation metrics provide useful insights into classifier performance, the sparse annotation nature of the HSI-Drive dataset images calls for analyzing the segmentation performance in detail by visualizing the segmentation of entire images.
This qualitative approach helps better understand the general performance of the segmentation system and its robustness, especially under challenging conditions.
In this section, we provide a summary of the quality analysis performed on several representative scenes.
Moreover, certain peculiarities of the system's operation, such as the segmentation of far scene backgrounds with low spatial resolution and severe spectral mixing, are better perceived through the analysis of video sequences rather than in still images.
The reader is referred to \cite{ipaccess} for the viewing of some example videos.
Since we have not yet explored any techniques to enhance ML model training using temporal information, these videos simply show the frame-to-frame segmentation generated by the FCNs.

\subsubsection{A highway scenario}
Fig. \ref{fig:557} depicts a highway scenario on a winter sunny morning, with traffic signs and guardrails on both sides, vegetation on one side, sky in the foreground, and vehicles about 25 meters ahead in both lanes.
Despite the challenging low-lighting conditions, the segmentation results are highly satisfactory.
In Experiment 3, the system effectively distinguishes the road signs, the coachwork of the cars (Painted Metal), and even detects the presence of a crane in the background of the image.

\subsubsection{Adverse lighting and weather conditions in highway}
Fig. \ref{fig:752} has been acquired on a winter rainy morning (two water droplets can be seen in the image).
It is important to note that for the Experiment 1, the FCN is robust in this situation and, for the Experiment 3 the segmentation errors occurs in the driving direction primarily because of the presence of some droplets.
Interestingly, when analyzing the video sequence corresponding to this image (refer to Fig. \ref{fig:summer_midday_wet_20000_AG2_f8}), it can be observed that the truck is correctly segmented in the frames prior to the appearance of the second droplet (frames 1 and 2).
Tarmac segmentation is robust in every moment despite the presence of the left droplet.

\subsubsection{Severe lighting contrasts}
The presence of shadows, particularly on sunny days, can result in significant lighting contrasts that challenge the dynamic range of the sensor and can hinder the accurate segmentation of scenes.
Fig. \ref{fig:560} and Fig. \ref{fig:228} illustrate two examples of this situation.
It can be observed that the FCN successfully prevents generating erroneous edges along the borders of the shadows, leading to an homogeneous segmentation where errors are mostly limited to small artifacts in the background.
In the image of Fig. \ref{fig:560}, captured on a sunny winter morning, even the small vehicles traveling in the opposite direction on the left side of the image are identified by the FCN.
However, due to the low resolution, it becomes challenging to distinguish between the coachwork and the lights of these vehicles.
Regarding Fig. \ref{fig:228}, despite two-thirds of the image being in the shade and only one-third in direct sunlight, there are no noticeable incorrect segmentations in Experiment 1.
In Experiment 3, there is only a small horizontal artifact of the Road Marks class produced by a speed bump.

\subsubsection{Overexposure}
The limited dynamic range of the sensor and the absence of automatic exposure control augment the likelihood of overexposure events, particularly under varying and high illumination conditions (reflections on surfaces, direct sunlight hitting the camera, etc.).
Pixel saturation can be catastrophic for the segmentation system, since the characteristic spectral signature of materials' reflectance is lost.
Fig. \ref{fig:566}, from a video recorded on a winter sunny morning, with frontal sunlight and severe glares on the tarmac, illustrates such a situation.
As can be seen, the scene is quite satisfactorily segmented as vehicles, tarmac, vegetation and even the guardrail are identified by the system.
The misclassified pixels are just some road marks erroneously classified as tarmac on the more overexposed sections.
To understand why this phenomenon does not more severely affect the overall segmentation, we show in Fig. \ref{fig:saturation} the significant differences in the number of saturated pixels across the 25 spectral bands.
The least saturated band (24) contains only 9124 saturated pixels, while the most saturated band (9), contains 21936 saturated pixels.
This demonstrates the advantage of using HSI with narrow, separated bands in tackling such situations.
In addition, it is interesting to note that even the most saturated band still provides valuable information (lights are clearly distinguished from the coachwork).

\subsubsection{Segmentation of scene backgrounds}
The low spatial resolution of the hyperspectral cubes challenges the accurate segmentation of objects in the background of images due to the lack of precise spatial information and the presence of strong spectral mixing.
However, this limitation does not significantly constraint the applicability of the system, since misidentified objects in the background are typically far away and they appear correctly segmented as the car moves forward and the distance to the object decreases.
An example of this can be observed in the sequence depicted in Fig. \ref{fig:winter_sunny_midday_10_AG1_f4}.
In the first frame of the sequence, a car is shown making a turn and moving downwards.
In that initial frame, a portion of the tarmac in the far background is incorrectly classified as either Vegetation or Other.
However, as the car moves forward in the subsequent frames, it can be observed that the same portion of tarmac is accurately segmented.

\subsubsection{Intra-class variability}
\begin{figure}[t]
\centering
\includegraphics[width=4cm]{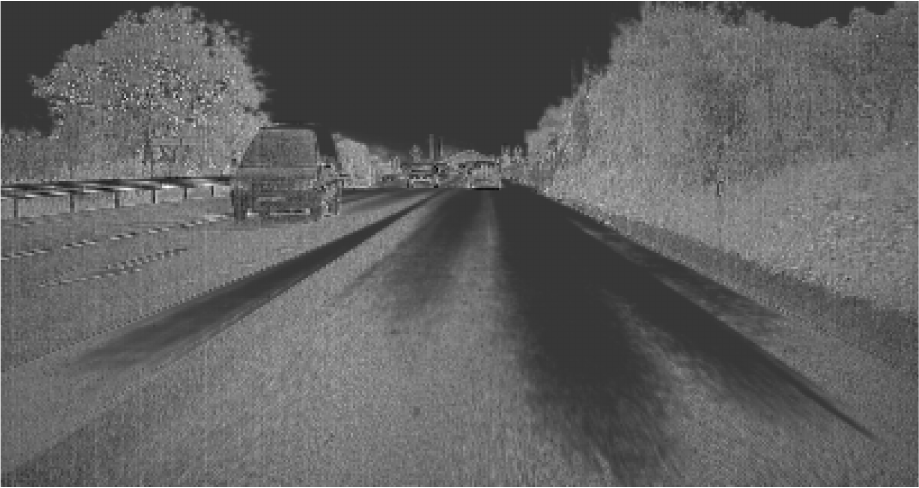}
\label{fig:mostSaturated}
\includegraphics[width=3.3cm]{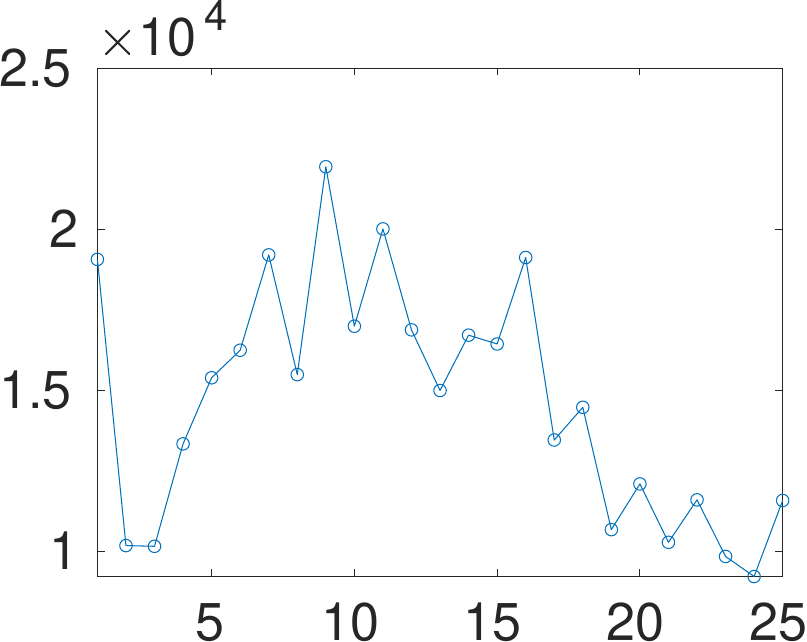}
\label{fig:saturatedPixelCount}
\includegraphics[width=4cm]{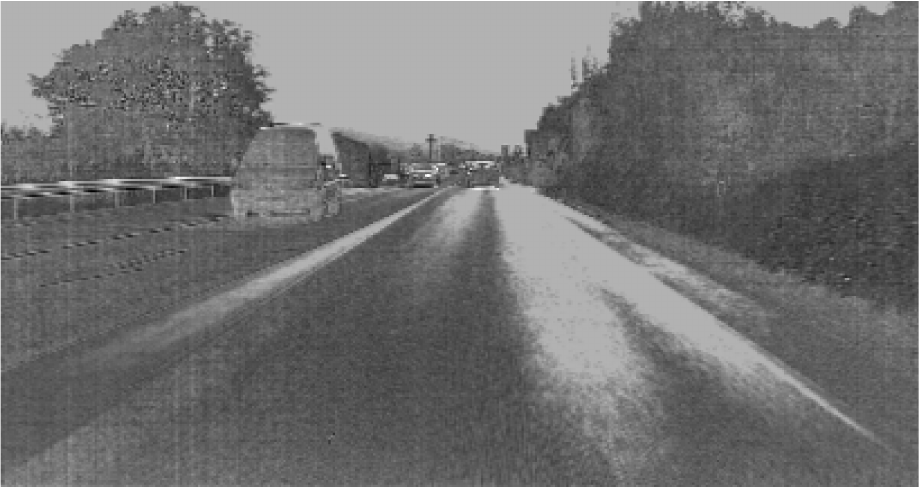}
\label{fig:minSaturated}
\caption{Grayscale images of the most saturated (left) and least saturated (right) bands and number of saturated pixels by band (center) of image 566, captured during a winter, sunny morning, in a road with overexposure}
\label{fig:saturation}

\centering
\includegraphics[width=5.5cm]{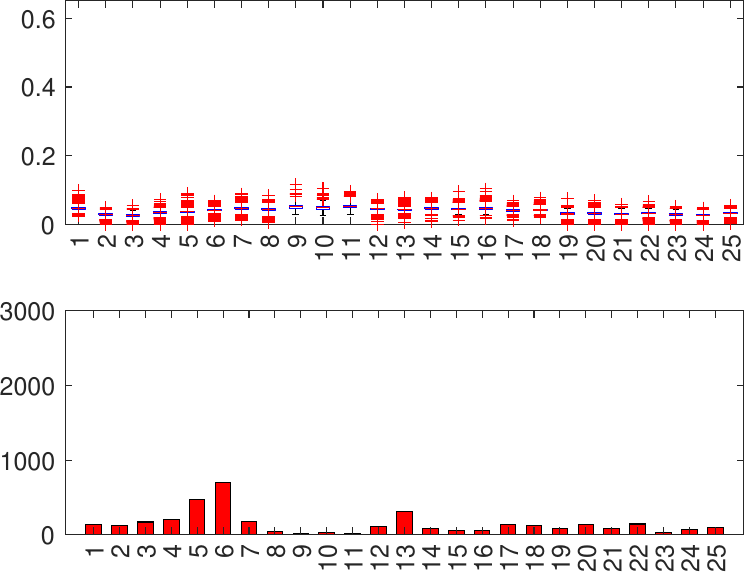}
\label{fig:RoadMarks}
\includegraphics[width=5.5cm]{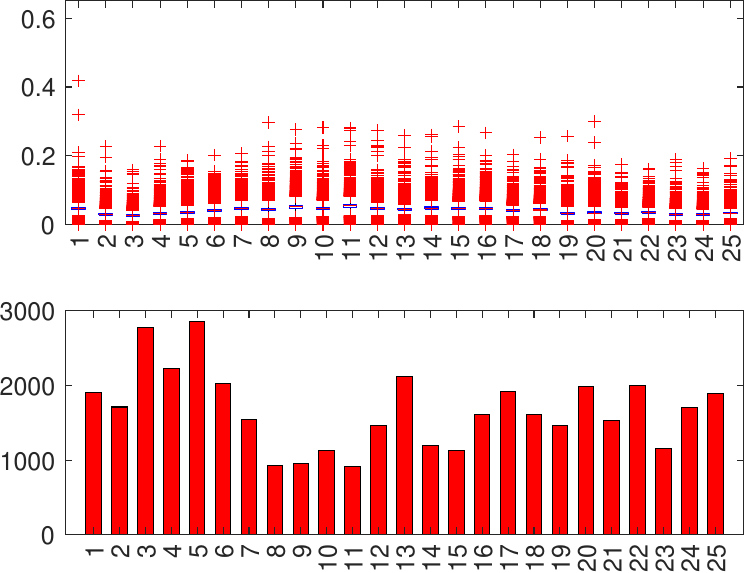}
\label{fig:PaintedMetal}
\includegraphics[width=5.5cm]{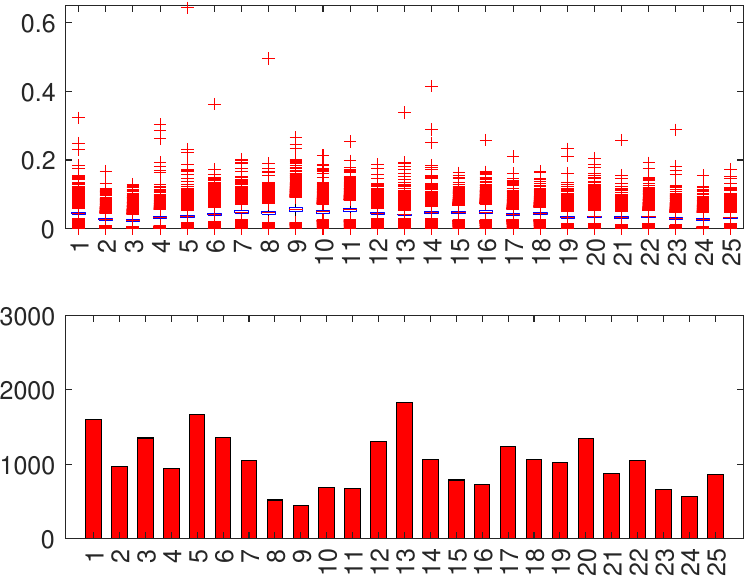}
\label{fig:Pedestrian}
\caption{Boxplots (top) and number of outliers (bottom) of Road Marks (left), Painted Metal (middle) and Pedestrian (right) classes using the spectral signatures of 100000 random pixels from each class.}
\label{fig:boxplot}

\centering
\includegraphics[width=2.75cm]{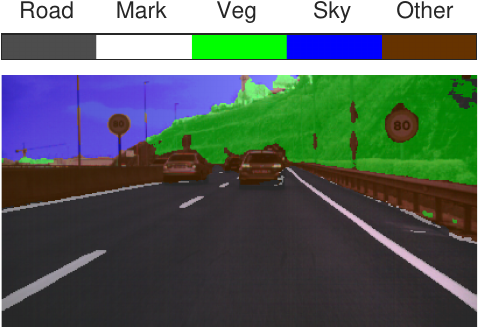}
\label{fig:557_1}
\includegraphics[width=2.75cm]{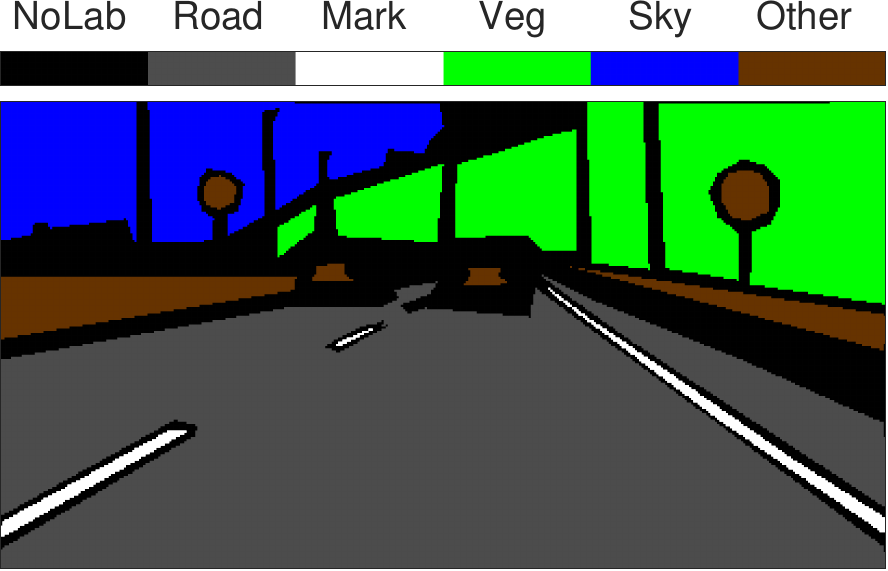}
\label{fig:557_2}
\includegraphics[width=2.75cm]{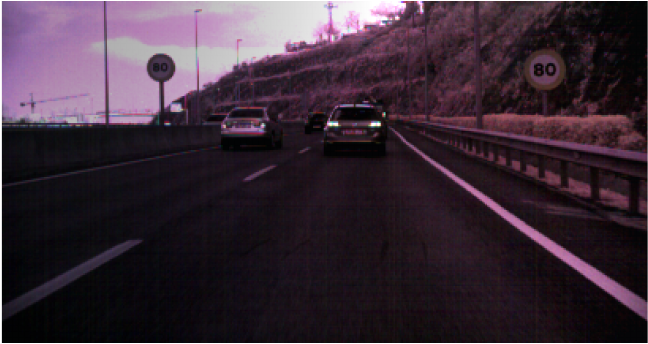}
\label{fig:557_3}
\includegraphics[width=2.75cm]{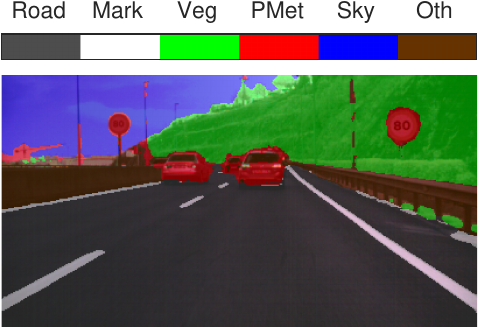}
\label{fig:557_4}
\includegraphics[width=2.75cm]{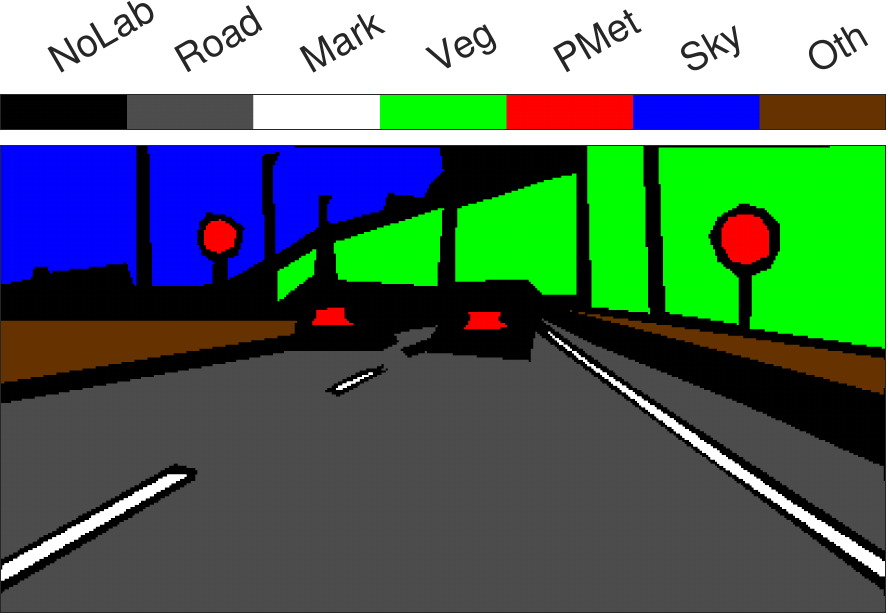}
\label{fig:557_5}
\caption{Image 557 (f4, AG2, 10ms), captured during a winter, sunny morning, in a highway: (far left) Exp2 segmentation, (left) Exp2 ground-truth, (center) false color, (right) Exp3 segmentation and (far right) Exp3 ground-truth.}
\label{fig:557}

\centering
\includegraphics[width=2.75cm]{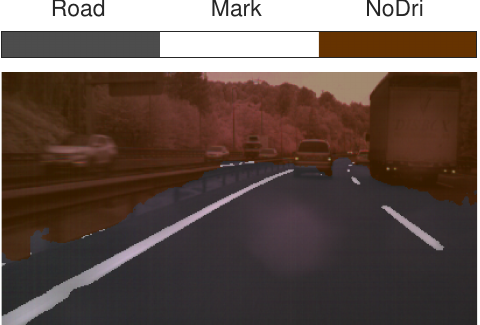}
\label{fig:752_1}
\includegraphics[width=2.75cm]{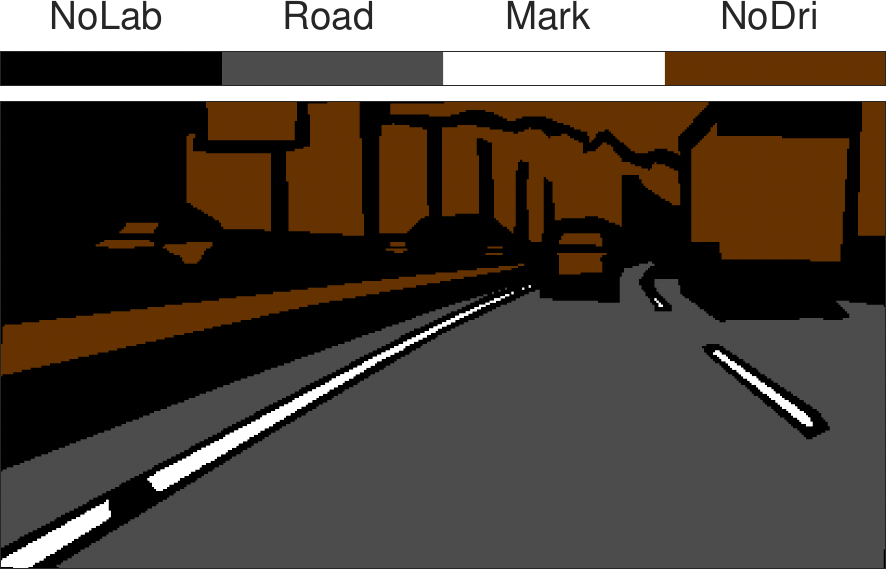}
\label{fig:752_2}
\includegraphics[width=2.75cm]{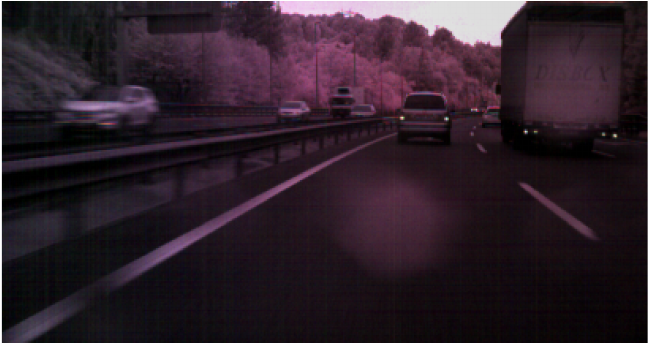}
\label{fig:752_3}
\includegraphics[width=2.75cm]{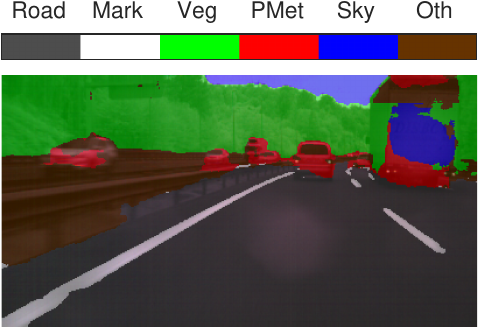}
\label{fig:752_4}
\includegraphics[width=2.75cm]{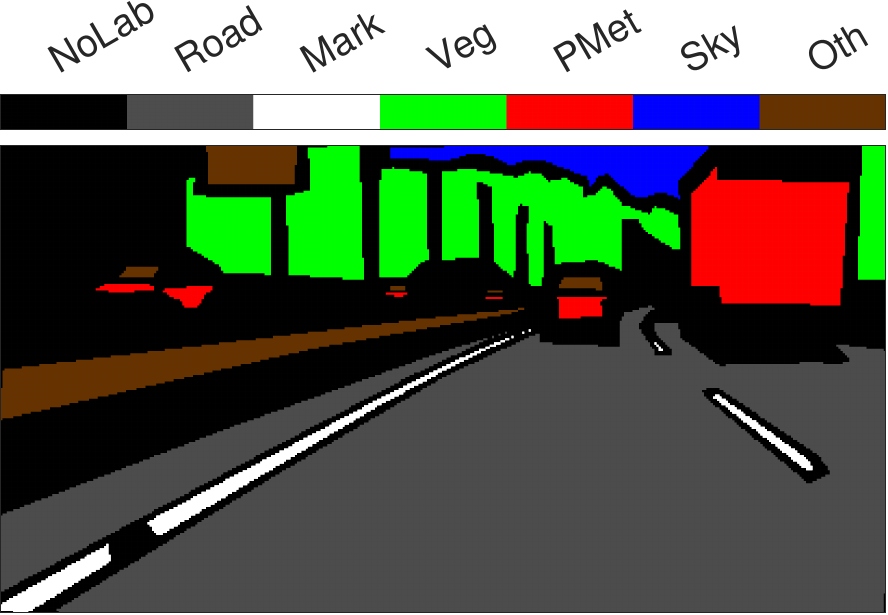}
\label{fig:752_5}
\caption{Image 752 (f8, AG2, 20ms), captured during a winter, rainy morning, in highway under adverse lighting and weather conditions: (far left) Exp1 segmentation, (left) Exp1 ground-truth, (center) false color, (right) Exp3 segmentation and (far right) Exp3 ground-truth.}
\label{fig:752}

\centering
\includegraphics[width=2.75cm]{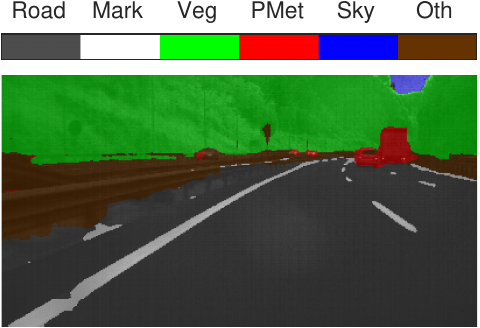}
\label{fig:summer_midday_wet_20000_AG2_f8_frame1}
\includegraphics[width=2.75cm]{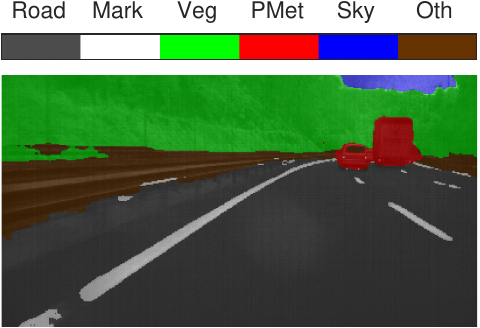}
\label{fig:summer_midday_wet_20000_AG2_f8_frame2}
\includegraphics[width=2.75cm]{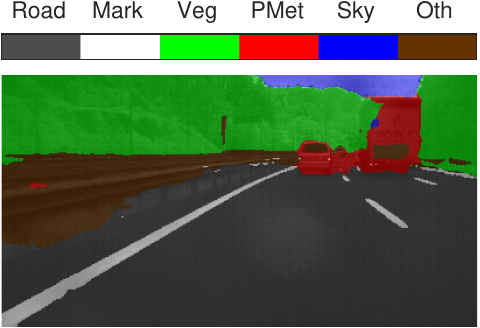}
\label{fig:summer_midday_wet_20000_AG2_f8_frame3}
\includegraphics[width=2.75cm]{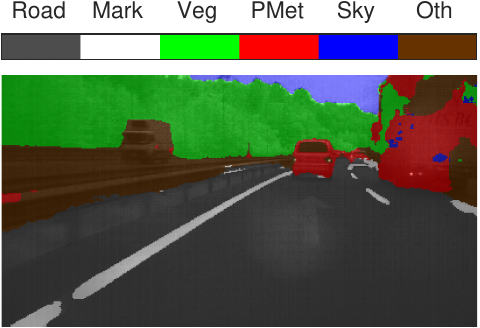}
\label{fig:summer_midday_wet_20000_AG2_f8_frame4}
\includegraphics[width=2.75cm]{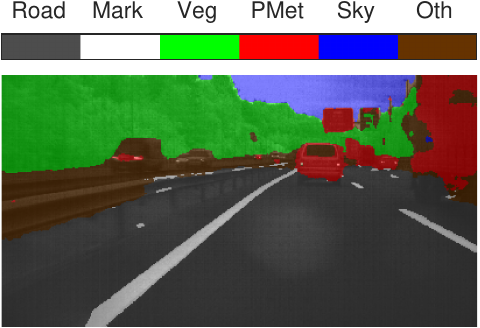}
\label{fig:summer_midday_wet_20000_AG2_f8_frame5}
\caption{Segmentation of video sequence 752, captured during a winter, rainy morning, in highway under adverse lighting and weather conditions. The time difference between every two frames is 3s.}
\label{fig:summer_midday_wet_20000_AG2_f8}

\centering
\includegraphics[width=2.75cm]{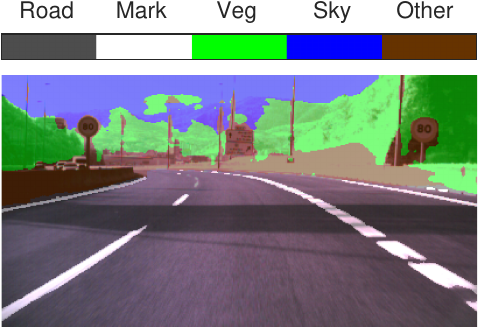}
\label{fig:560_1}
\includegraphics[width=2.75cm]{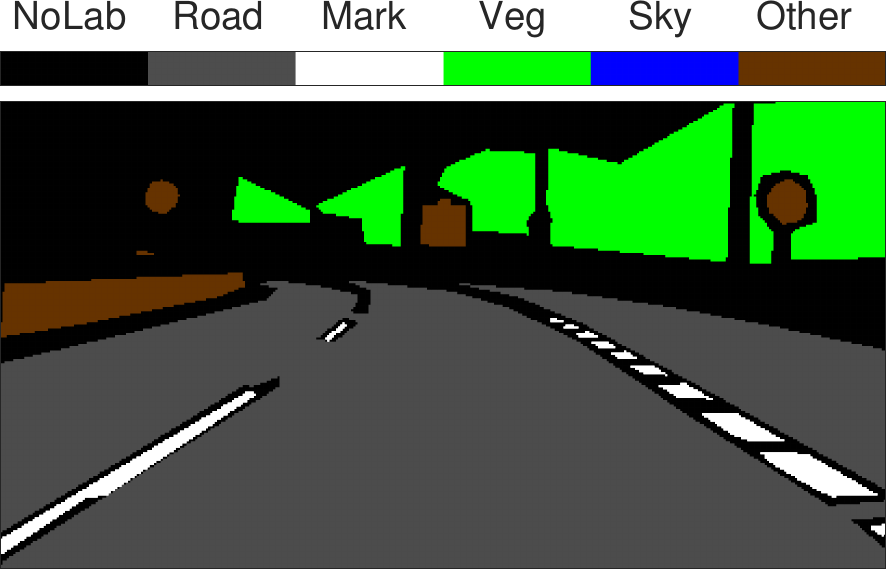}
\label{fig:560_2}
\includegraphics[width=2.75cm]{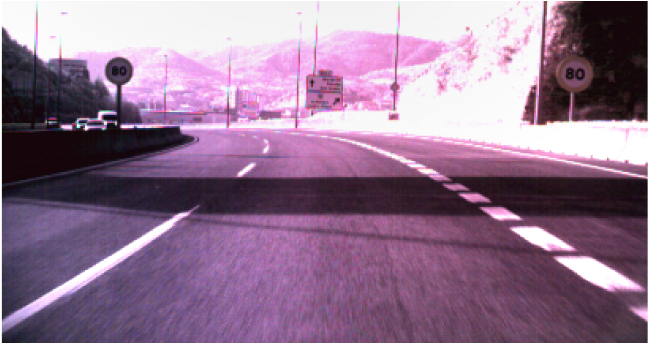}
\label{fig:560_3}
\includegraphics[width=2.75cm]{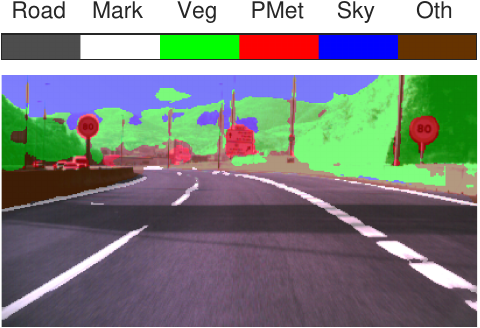}
\label{fig:560_4}
\includegraphics[width=2.75cm]{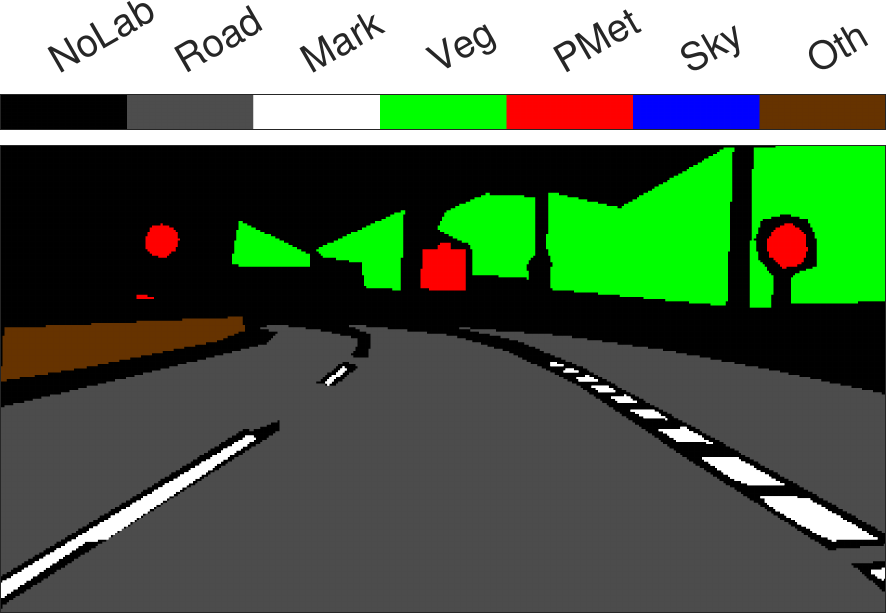}
\label{fig:560_5}
\caption{Image 560 (f4, AG2, 10ms), captured during a winter, sunny morning, in road with intense contrasts: (far left) Exp2 segmentation, (left) Exp2 ground-truth, (center) false color, (right) Exp3 segmentation and (far right) Exp3 ground-truth.}
\label{fig:560}

\centering
\includegraphics[width=2.75cm]{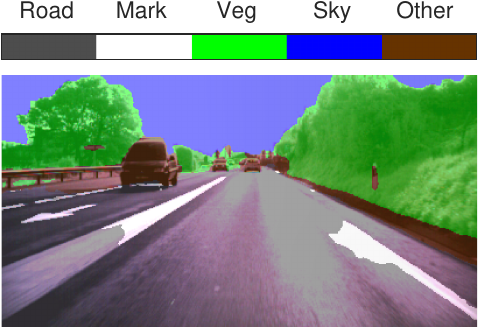}
\label{fig:566_1}
\includegraphics[width=2.75cm]{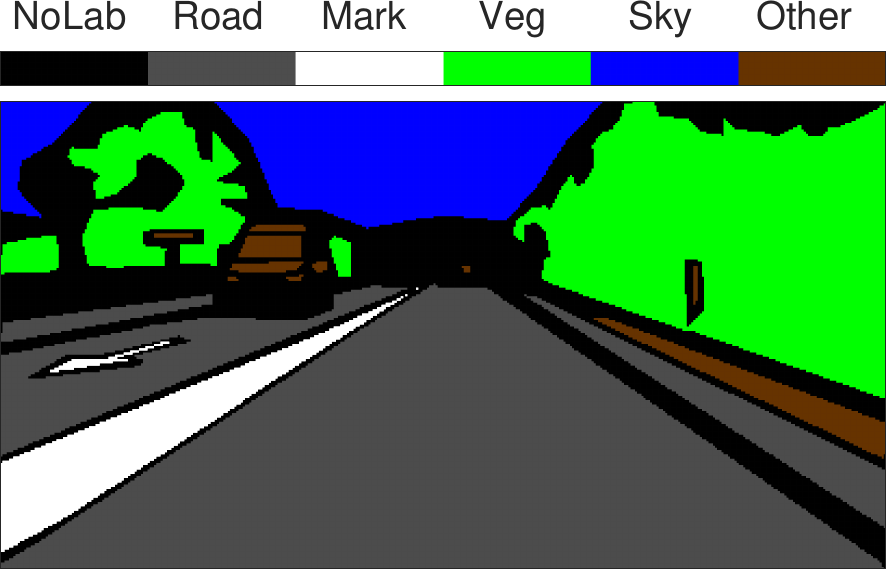}
\label{fig:566_2}
\includegraphics[width=2.75cm]{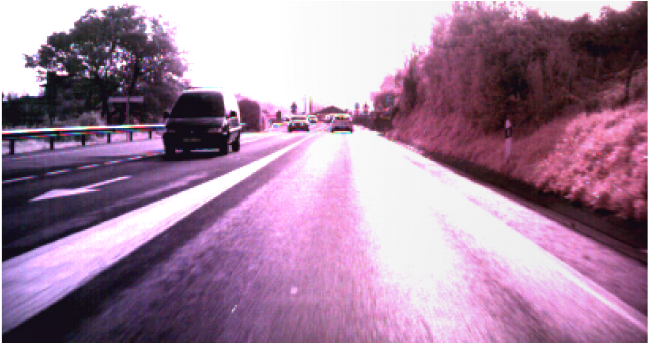}
\label{fig:566_3}
\includegraphics[width=2.75cm]{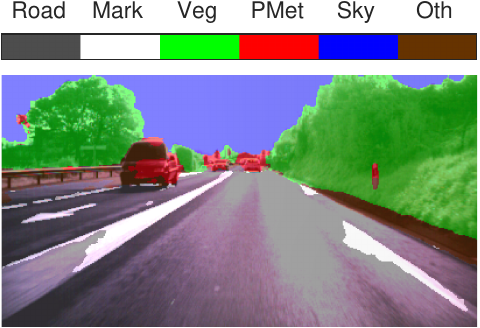}
\label{fig:566_4}
\includegraphics[width=2.75cm]{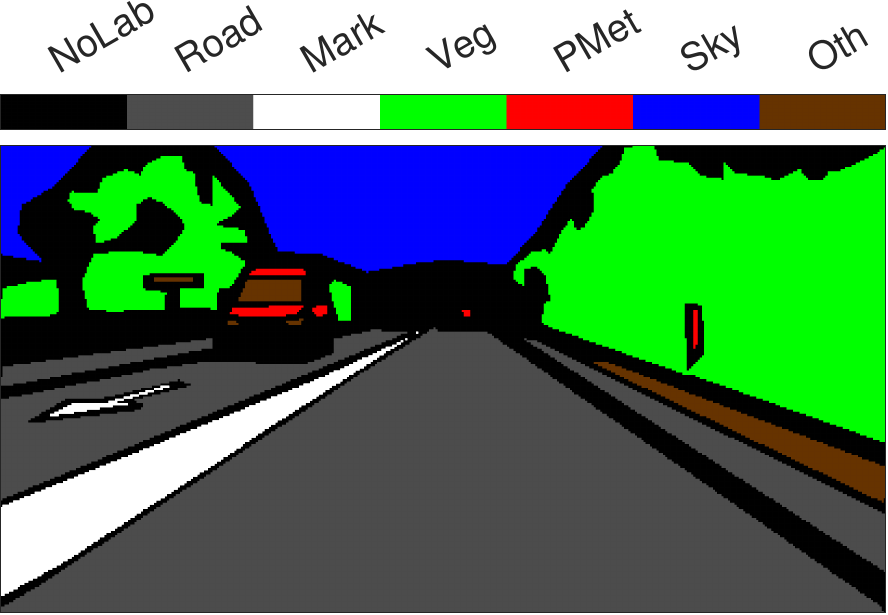}
\label{fig:566_5}
\caption{Image 566 (f4, AG2, 10ms), captured during a winter, sunny morning, in a road with overexposure: (far left) Exp2 segmentation, (left) Exp2 ground-truth, (center) false color, (right) Exp3 segmentation and (far right) Exp3 ground-truth).}
\label{fig:566}
\end{figure}

The Painted Metal category, as an example, contains various object-types such as speed signs, information panels, vehicles, traffic lights or street lamps.
Similarly, the Pedestrian class encompasses pedestrians, cyclists, motorcyclists and even animals, where the differences in clothing further contribute to the spectral diversity.
As mentioned in Section \ref{sec:experimentalsetup}, the high intra-class spectral variability of these classes can be a handicap for their correct classification.
To better illustrate the difference between these classes and other classes with low variability, Fig. \ref{fig:boxplot} shows box plots and histograms of outliers in the spectral signatures of 100,000 random pixels from three minority classes: Road Marks, Painted Metal, and Pedestrian.
It can be observed that Road Marks exhibits a more compact distribution with fewer outliers compared to the other two classes.
Painted Metal and Pedestrian exhibit alternate variability across the spectral bands, but Painted Metal contains more outliers in each band.
These findings align with the numerical results presented in Tables \ref{tab:exp1} to \ref{tab:exp4}.

\begin{figure}[t]
\centering
\includegraphics[width=2.75cm]{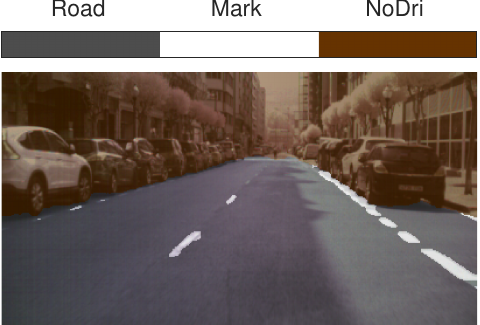}
\label{fig:228_1}
\includegraphics[width=2.75cm]{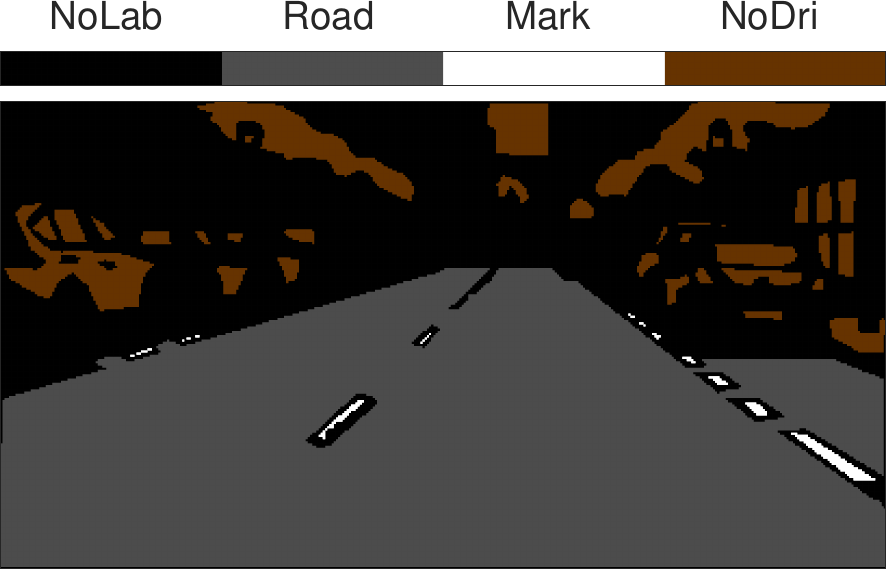}
\label{fig:228_2}
\includegraphics[width=2.75cm]{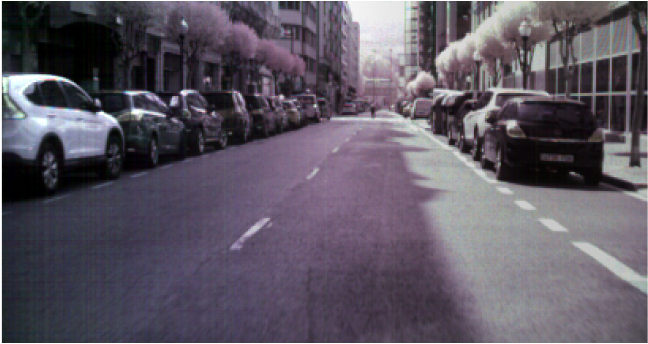}
\label{fig:228_3}
\includegraphics[width=2.75cm]{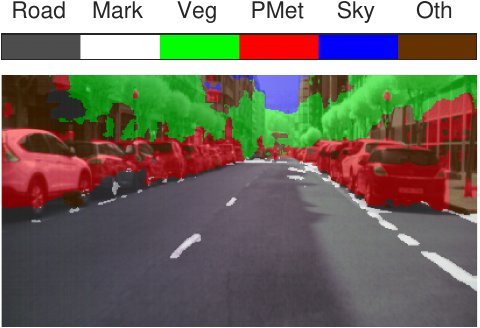}
\label{fig:228_4}
\includegraphics[width=2.75cm]{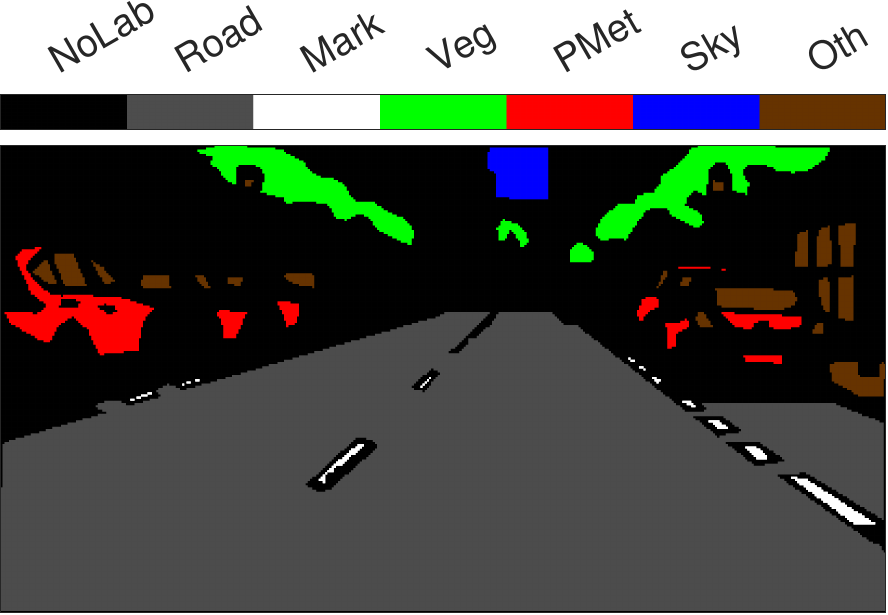}
\label{fig:228_5}
\caption{Image 228 (f8, AG1, 10ms), captured during a spring, sunny midday, in an urban environment with shadows: (far left) Exp1 segmentation, (left) Exp1 ground-truth, (center) false color, (right) Exp3 segmentation and (right) Exp3 ground-truth.}
\label{fig:228}

\centering
\includegraphics[width=2.75cm]{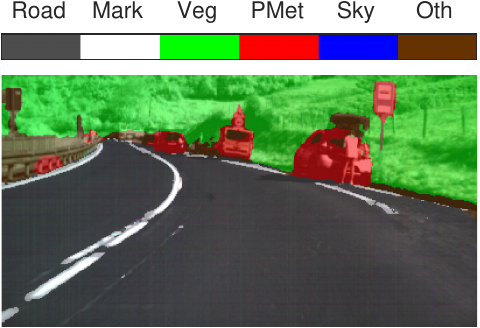}
\label{fig:404_1}
\includegraphics[width=2.75cm]{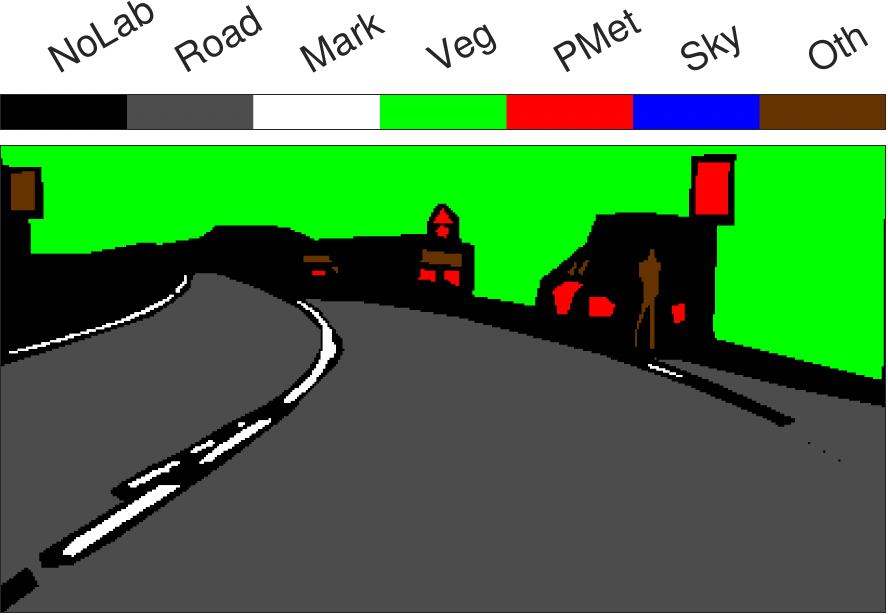}
\label{fig:404_2}
\includegraphics[width=2.75cm]{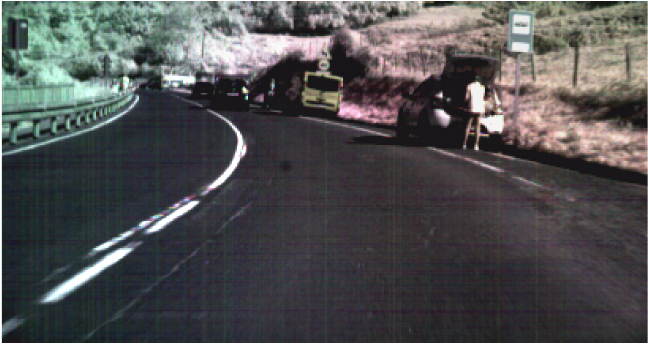}
\label{fig:404_3}
\includegraphics[width=2.75cm]{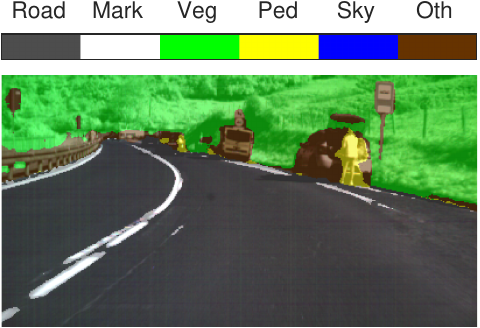}
\label{fig:404_4}
\includegraphics[width=2.75cm]{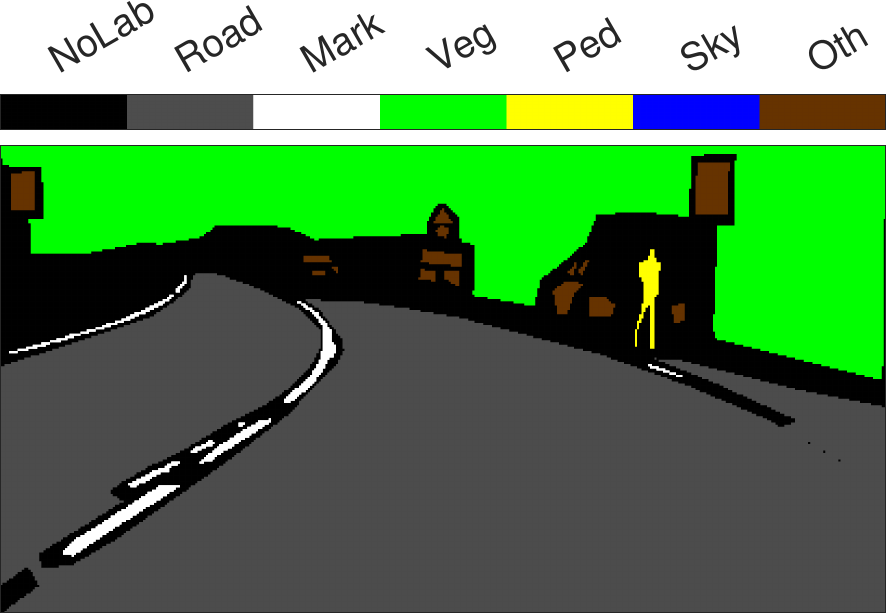}
\label{fig:404_5}
\caption{Image 404 (f8, AG1, 10ms), captured during a fall, sunny midday, in road with Painted/Unpainted metal objects with similar shape: (far left) Exp3 segmentation, (left) Exp3 ground-truth, (center) false color, (right) Exp4 segmentation and (far right) Exp4 ground-truth.}
\label{fig:404}

\centering
\includegraphics[width=2.75cm]{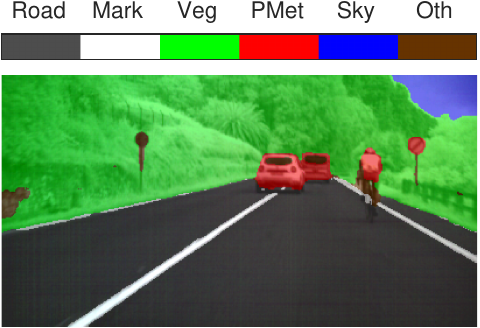}
\label{fig:301_1}
\includegraphics[width=2.75cm]{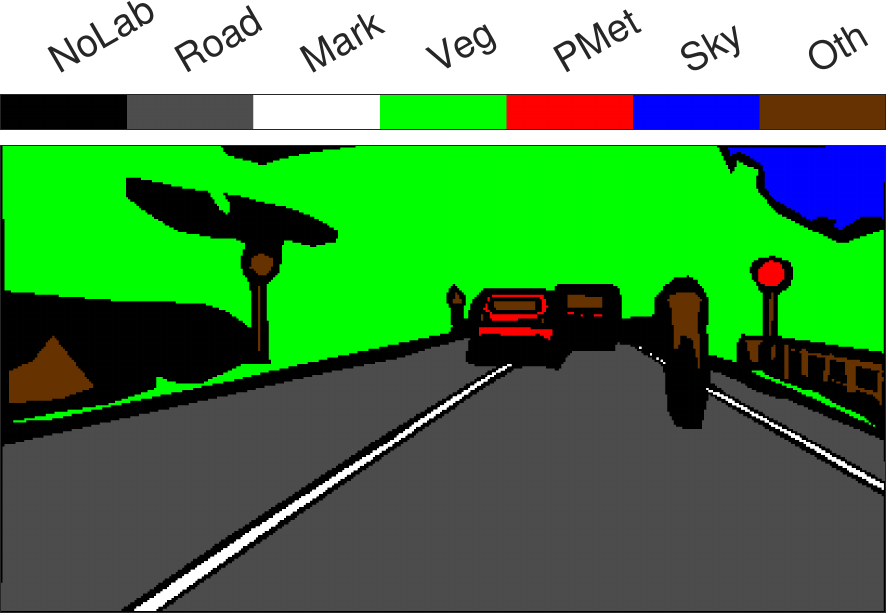}
\label{fig:301_2}
\includegraphics[width=2.75cm]{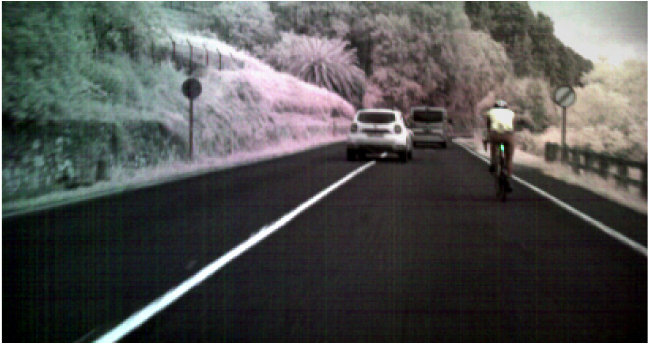}
\label{fig:301_3}
\includegraphics[width=2.75cm]{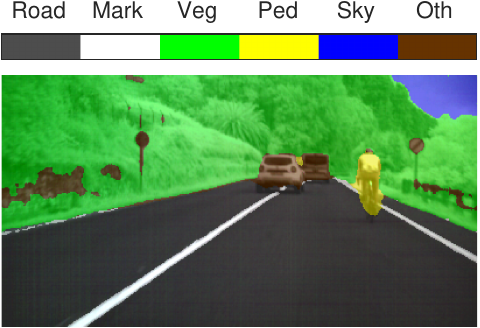}
\label{fig:301_4}
\includegraphics[width=2.75cm]{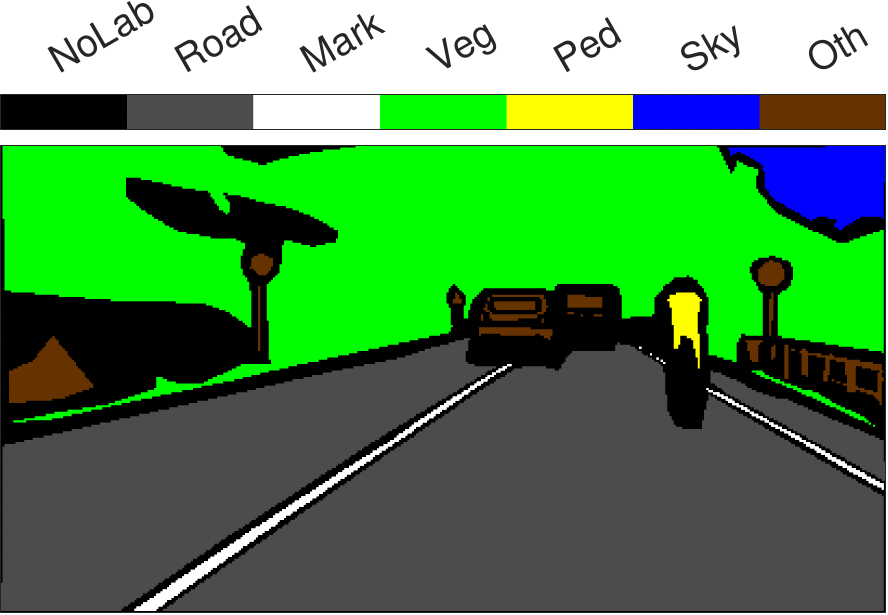}
\label{fig:301_5}
\caption{Image 301 (f8, AG2, 20ms), captured during a summer, rainy morning, with a cyclist and Painted Metal objects: (far left) Exp3 segmentation, (left) Exp3 ground-truth, (center) false color, (right) Exp4 segmentation and (far right) Exp4 ground-truth.}
\label{fig:301}

\centering
\includegraphics[width=2.75cm]{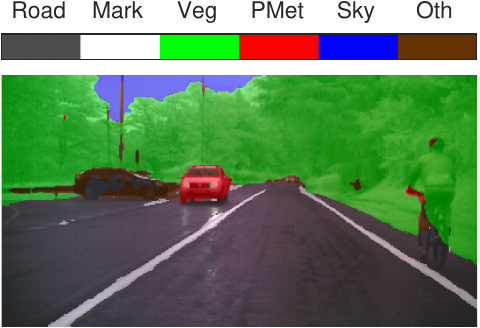}
\label{fig:677_1}
\includegraphics[width=2.75cm]{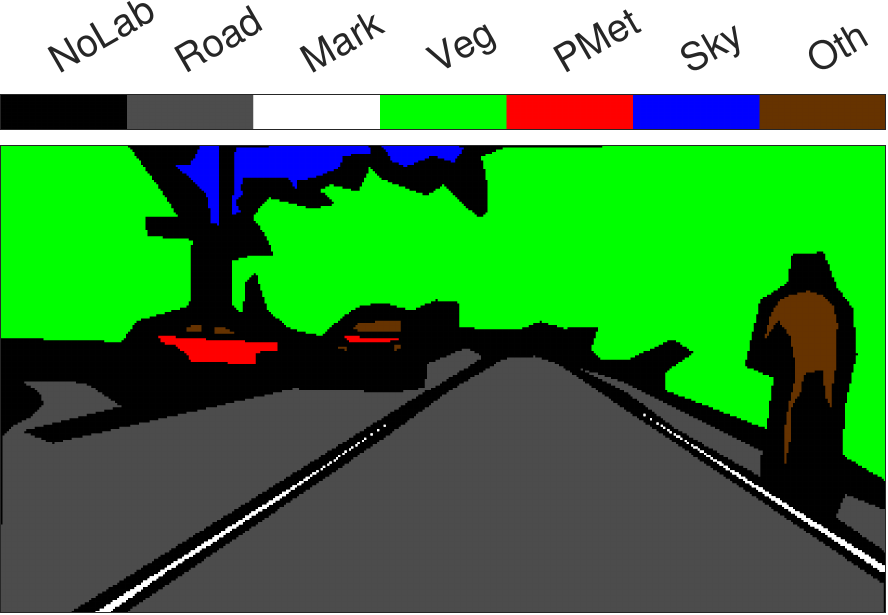}
\label{fig:677_2}
\includegraphics[width=2.75cm]{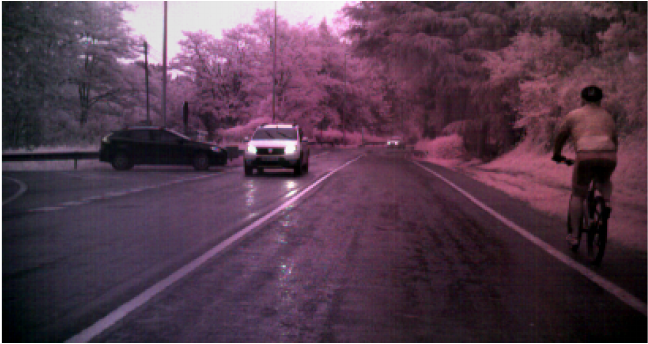}
\label{fig:677_3}
\includegraphics[width=2.75cm]{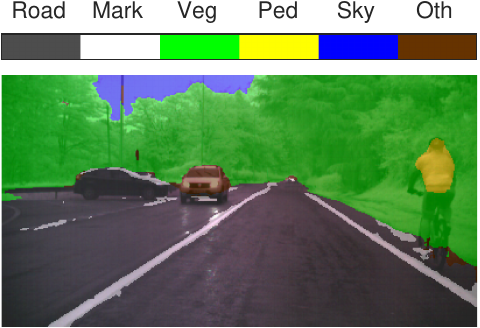}
\label{fig:677_4}
\includegraphics[width=2.75cm]{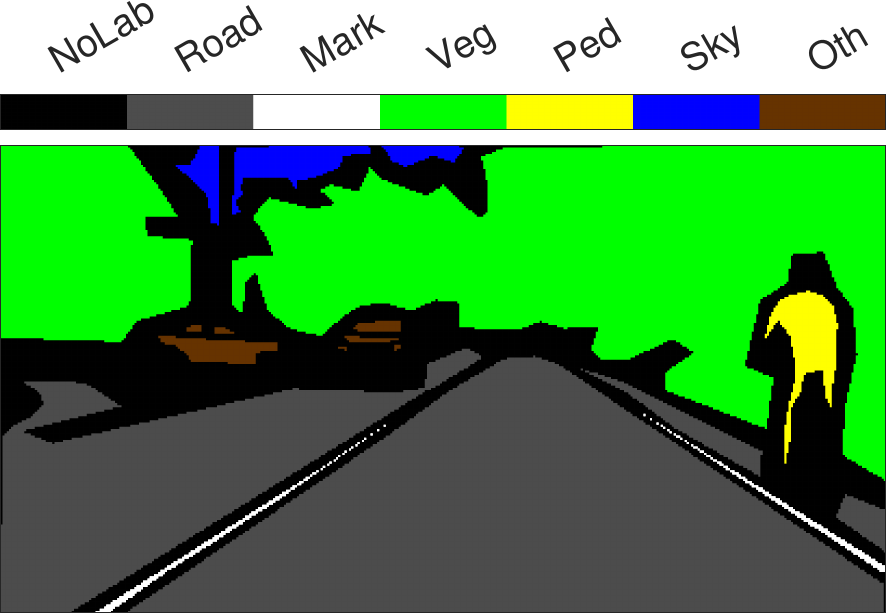}
\label{fig:677_5}
\caption{Image 677 (f4, AG2, 10ms), captured during a spring, rainy midday, in a road with a cyclist on the right shoulder: (far left) Exp3 segmentation, (left) Exp3 ground-truth, (visible) false color, (right) Exp4 segmentation and (far right) Exp4 ground-truth.}
\label{fig:677}

\centering
\includegraphics[width=2.75cm]{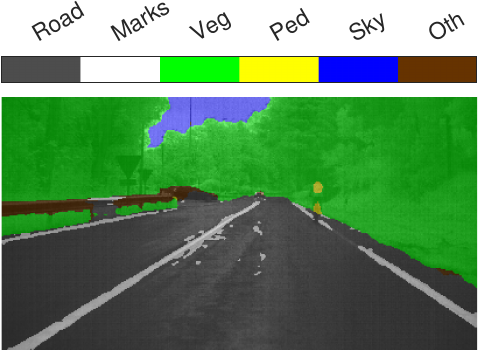}
\label{fig:spring_wet_midday_10000_AG2_f4_frame1}
\includegraphics[width=2.75cm]{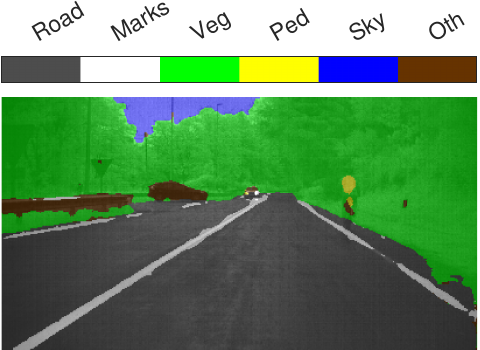}
\label{fig:spring_wet_midday_10000_AG2_f4_frame2}
\includegraphics[width=2.75cm]{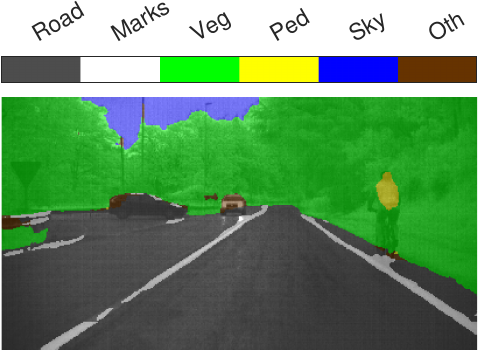}
\label{fig:spring_wet_midday_10000_AG2_f4_frame3}
\includegraphics[width=2.75cm]{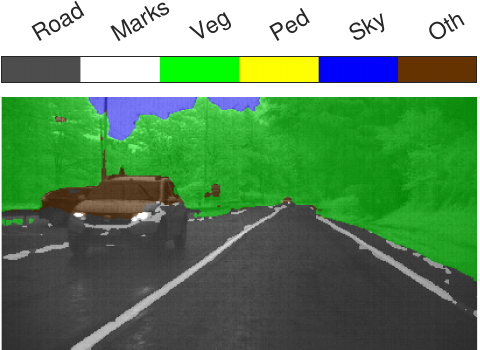}
\label{fig:spring_wet_midday_10000_AG2_f4_frame4}
\includegraphics[width=2.75cm]{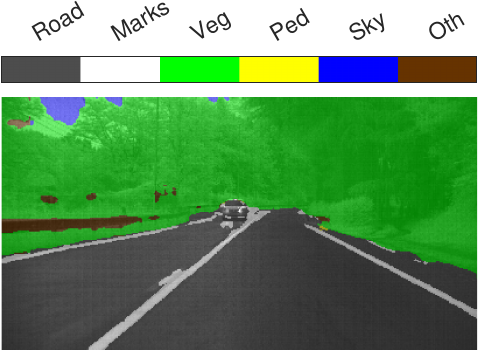}
\label{fig:spring_wet_midday_10000_AG2_f4_frame5}
\caption{Segmentation of video sequence 677, captured during a spring, rainy midday, in a road with a cyclist on the right shoulder. The time difference between every two frames is 2s.}
\label{fig:spring_wet_midday_10000_AG2_f4}

\centering
\includegraphics[width=2.75cm]{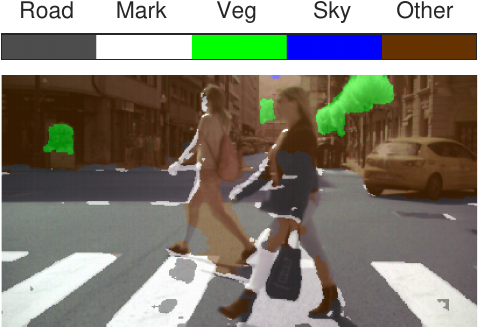}
\label{fig:229_1}
\includegraphics[width=2.75cm]{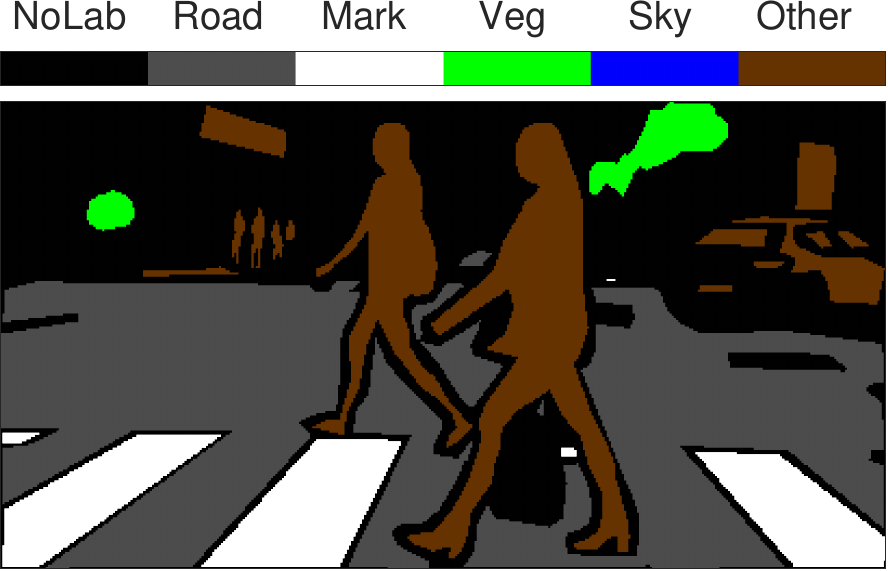}
\label{fig:229_2}
\includegraphics[width=2.75cm]{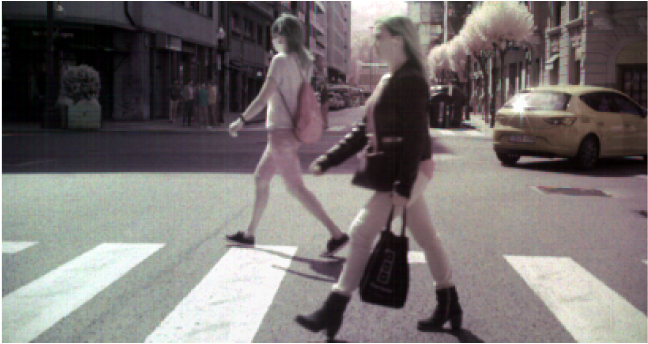}
\label{fig:229_3}
\includegraphics[width=2.75cm]{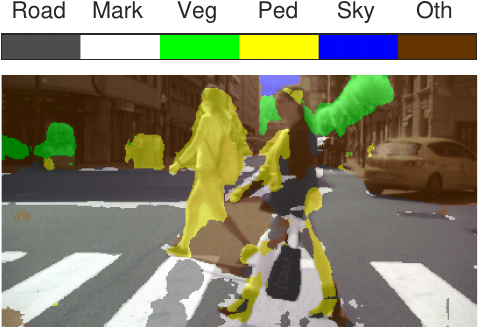}
\label{fig:229_4}
\includegraphics[width=2.75cm]{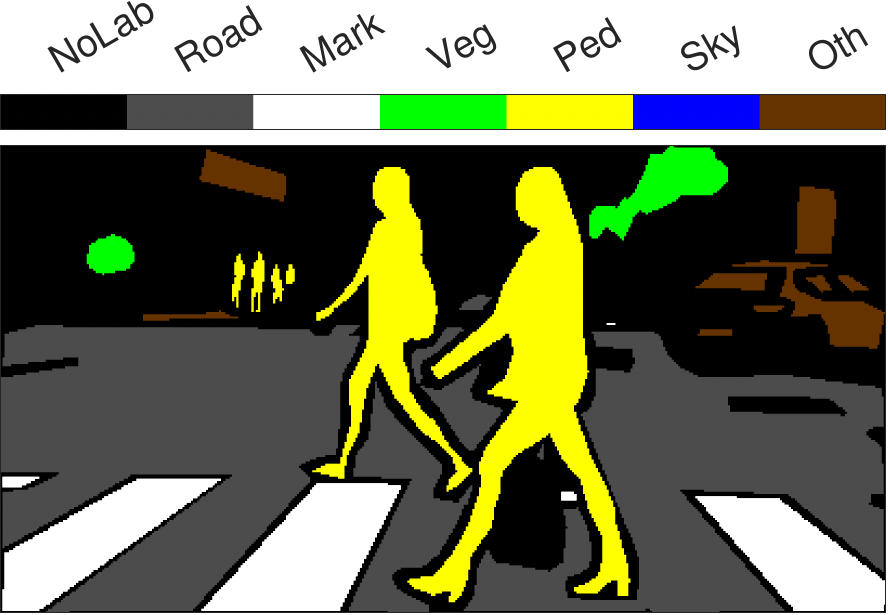}
\label{fig:229_5}
\caption{Image 229 (f8, AG1, 10ms), captured during a spring, sunny midday, in urban environment with two pedestrians in a zebra crossing: (far left) Exp2 segmentation, (left) Exp2 ground-truth, (center) false color, (right) Exp4 segmentation and (far right) Exp4 ground-truth.}
\label{fig:229}

\centering
\includegraphics[width=2.75cm]{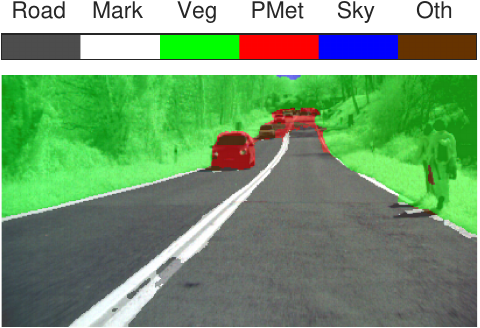}
\label{fig:651_1}
\includegraphics[width=2.75cm]{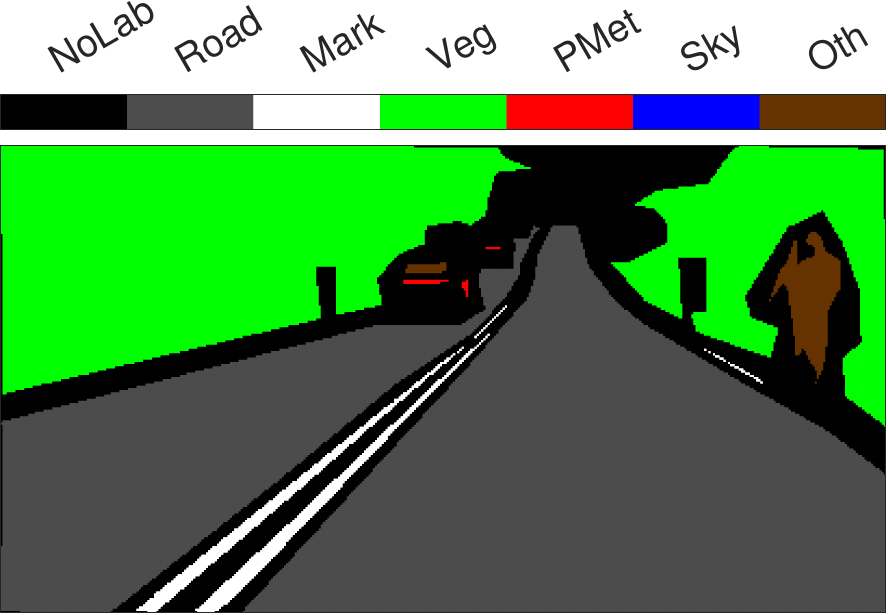}
\label{fig:651_2}
\includegraphics[width=2.75cm]{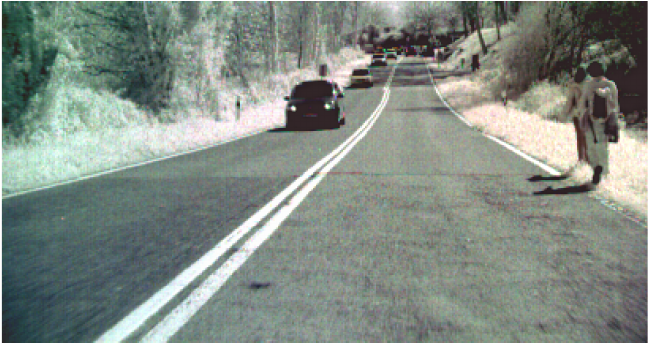}
\label{fig:651_3}
\includegraphics[width=2.75cm]{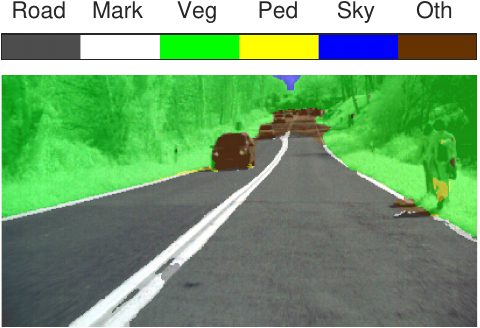}
\label{fig:651_4}
\includegraphics[width=2.75cm]{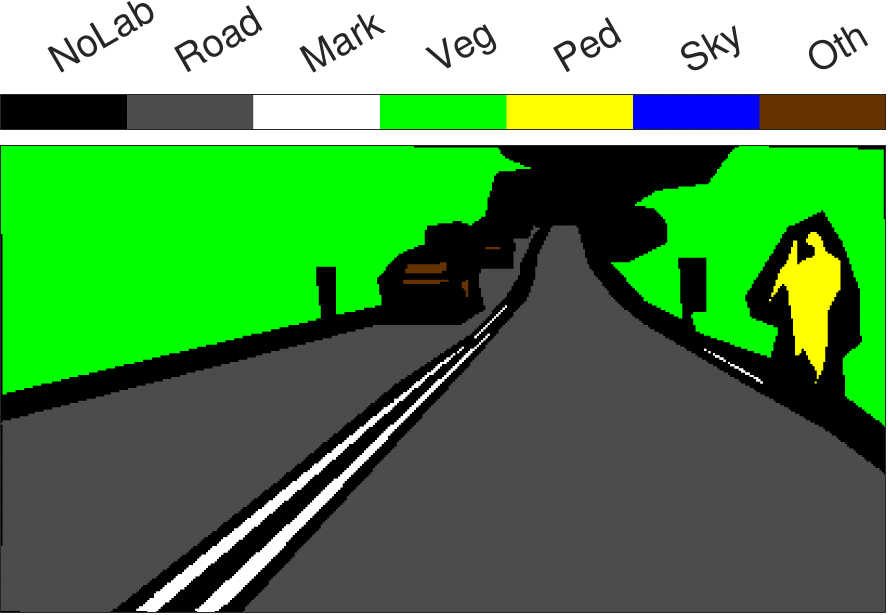}
\label{fig:651_5}
\caption{Image 651 (f4, AG1, 10ms), captured during a winter, sunny midday, in road with two pedestrians walking in the right road shoulder: (far left) Exp3 segmentation, (left) Exp3 ground-truth, (center) false color, (right) Exp4 segmentation and (far right) Exp4 ground-truth.}
\label{fig:651}

\centering
\includegraphics[width=2.7cm]{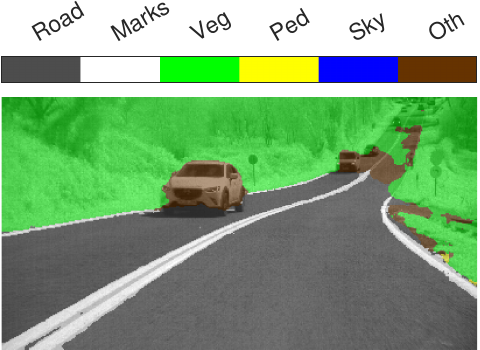}
\label{fig:winter_sunny_midday_10_AG1_f4_frame1}
\includegraphics[width=2.7cm]{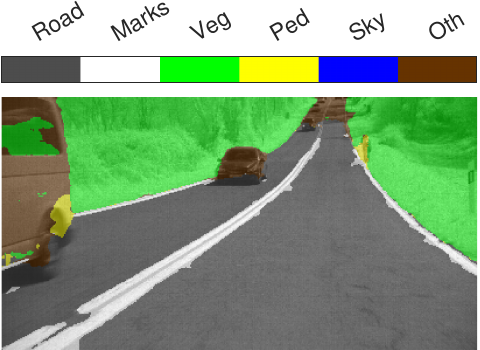}
\label{fig:winter_sunny_midday_10_AG1_f4_frame2}
\includegraphics[width=2.7cm]{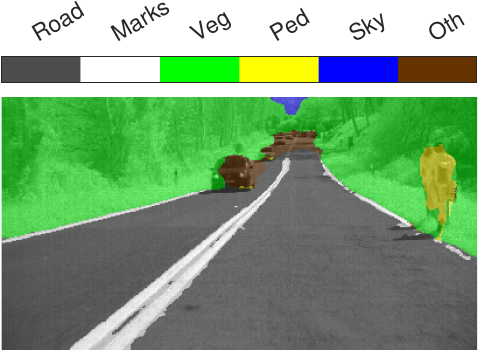}
\label{fig:winter_sunny_midday_10_AG1_f4_frame3}
\includegraphics[width=2.7cm]{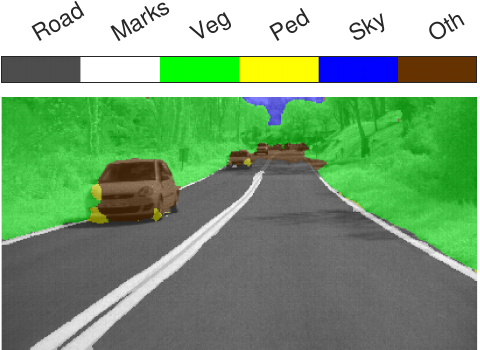}
\label{fig:winter_sunny_midday_10_AG1_f4_frame4}
\includegraphics[width=2.7cm]{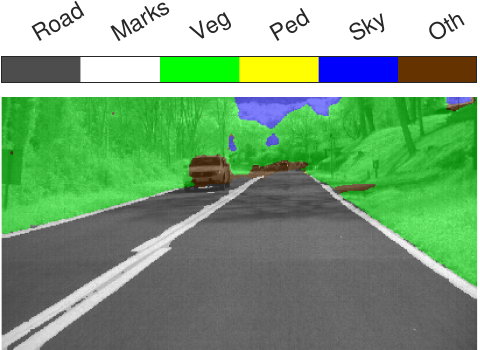}
\label{fig:winter_sunny_midday_10_AG1_f4_frame5}
\caption{Segmentation of video sequence 651, captured during a winter, sunny midday, in road with two pedestrians walking in the right road shoulder. The time difference between every two frames is 2s.}
\label{fig:winter_sunny_midday_10_AG1_f4}
\end{figure}

\paragraph{Painted Metal}
Despite its high intra-class variability, the significant contribution of spectral information to the segmentation performance becomes evident if one observes, for example, how the FCN correctly differentiates the front views (Painted Metal) and the rear views (Unpainted Metal/Other) of signals as shown in Fig. \ref{fig:404} and in Fig. \ref{fig:301}.
However, there are also situations where the network is not robust, such as that shown in Fig. \ref{fig:677}, where a black-painted vehicle is sometimes confused with tarmac.
Although good segmentation of dark vehicles has also been obtained in other situatoins (see Fig. \ref{fig:557}, Fig. \ref{fig:566}, Fig. \ref{fig:228}, and Fig. \ref{fig:651}), there is no clear evidence of the metamerism of RGB images being completely overcome in this case.

\paragraph{Pedestrian}
Hereunder, we show examples of how the spectrally diverse elements that comprise Pedestrian class are segmented.
In Fig. \ref{fig:404}, the correct identification of a pedestrian on the road shoulder can be observed.
Fig. \ref{fig:301} and Fig. \ref{fig:677} provide two good examples where a cyclist on the right road shoulder is quite accurately detected in an interurban road on a rainy morning.
However, in Fig. \ref{fig:229}, although the pedestrians in the background and the woman in the second plane are correctly identified, the FCN is not able to detect the woman in the foreground.
Similarly, in Fig. \ref{fig:651} a couple of pedestrians have not been identified by the FCN.
Nevertheless, when we examine frames from the corresponding video sequence  (Fig. \ref{fig:winter_sunny_midday_10_AG1_f4}), we can observe how, in the second frame, the pedestrians are detected even when they are far away and, in the third frame, they are accurately segmented.
Further investigation is needed to understand the cause of this instability and improve the overall performance of pedestrian segmentation.

\section{Conclusions}\label{sec:conclu}
This article introduces HSI-Drive v2.0, the second version of the HSI-Drive dataset, comprising 752 images depicting real traffic scenarios throughout all seasons of the year.
The dataset contains approximately 44 million manually labeled pixels divided into 10 categories, based predominantly on the spectral reflectance properties of materials found in driving environments.
This extended dataset significantly augments the pixel count for the underrepresented classes, which enables the development of more accurate and robust ML segmentation models for improved scene understanding in ADS.

The potential of this new dataset is demonstrated through various experiments with a newly redesigned FCN model, showcasing substantial improvements over previous results obtained with version v1.1.
The updated model has also been evaluated in two new six-class experiments comprising the Painted Metal and Pedestrian classes.
Despite the high spectral intra-class variability in these classes, the results remain quite satisfactory, considering that the model was trained and tested on data captured under highly variable and challenging lighting and weather conditions.

Future work will concentrate on enhancing the segmentation system's overall performance in two key directions.
First, the adoption of edge preserving techniques will be explored to achieve more accurate object-background boundaries.
Secondly, spatio-temporal approaches will be essayed not only to improve video segmentation accuracy but also to reduce the computational load of sequential frame-to-frame segmentation.
Additionally, further investigation will be conducted to better understand the contribution of hyperspectral information in overcoming the metamerism of RGB imaging, specially under challenging conditions.
Finally, the models and algorithms will have to be optimized and efficient and secure processing architectures developed, to enable the deployment of these systems on resource and power constrained embedded platforms suitable for the implementation of ADAS and ADS.
\bibliographystyle{IEEEtran}
\bibliography{IEEEabrv, mybiblio.bib}

\begin{thebibliography}{10}
\providecommand{\url}[1]{#1}
\csname url@samestyle\endcsname
\providecommand{\newblock}{\relax}
\providecommand{\bibinfo}[2]{#2}
\providecommand{\BIBentrySTDinterwordspacing}{\spaceskip=0pt\relax}
\providecommand{\BIBentryALTinterwordstretchfactor}{4}
\providecommand{\BIBentryALTinterwordspacing}{\spaceskip=\fontdimen2\font plus
\BIBentryALTinterwordstretchfactor\fontdimen3\font minus
  \fontdimen4\font\relax}
\providecommand{\BIBforeignlanguage}[2]{{%
\expandafter\ifx\csname l@#1\endcsname\relax
\typeout{** WARNING: IEEEtran.bst: No hyphenation pattern has been}%
\typeout{** loaded for the language `#1'. Using the pattern for}%
\typeout{** the default language instead.}%
\else
\language=\csname l@#1\endcsname
\fi
#2}}
\providecommand{\BIBdecl}{\relax}
\BIBdecl

\bibitem{basterretxeahsi}
K.~Basterretxea, V.~Martínez, J.~Echanobe, J.~Gutiérrez–Zaballa, and
  I.~Del~Campo, ``{HSI-Drive: A Dataset for the Research of Hyperspectral Image
  Processing Applied to Autonomous Driving Systems},'' in \emph{2021 IEEE
  Intelligent Vehicles Symposium (IV)}, 2021, pp. 866--873.

\bibitem{mv1-d2048x1088-hs02-96-g2}
\BIBentryALTinterwordspacing
Photonfocus, ``{MV1-D2048x1088-HS02-96-G2}.'' [Online]. Available:
  \url{https://www.photonfocus.com/products/camerafinder/camera/mv1-d2048x1088-hs02-96-g2}
\BIBentrySTDinterwordspacing

\bibitem{phan2022efficient}
M.~H. Phan, S.~L. Phung, K.~Luu, and A.~Bouzerdoum, ``{Efficient hyperspectral
  image segmentation for biosecurity scanning using knowledge distillation from
  multi-head teacher},'' \emph{Neurocomputing}, vol. 504, pp. 189--203, 2022.

\bibitem{taghizadeh2011comparison}
M.~Taghizadeh, A.~A. Gowen, and C.~P. O’Donnell, ``{Comparison of
  hyperspectral imaging with conventional RGB imaging for quality evaluation of
  Agaricus bisporus mushrooms},'' \emph{Biosystems engineering}, vol. 108,
  no.~2, pp. 191--194, 2011.

\bibitem{seidlitz2022robust}
S.~Seidlitz, J.~Sellner, J.~Odenthal, B.~{\"O}zdemir, A.~Studier-Fischer,
  S.~Kn{\"o}dler, L.~Ayala, T.~J. Adler, H.~G. Kenngott, M.~Tizabi
  \emph{et~al.}, ``{Robust deep learning-based semantic organ segmentation in
  hyperspectral images},'' \emph{Medical Image Analysis}, p. 102488, 2022.

\bibitem{fricker2019convolutional}
G.~A. Fricker, J.~D. Ventura, J.~A. Wolf, M.~P. North, F.~W. Davis, and
  J.~Franklin, ``{A convolutional neural network classifier identifies tree
  species in mixed-conifer forest from hyperspectral imagery},'' \emph{Remote
  Sensing}, vol.~11, no.~19, p. 2326, 2019.

\bibitem{winkens2017hyko}
C.~Winkens, F.~Sattler, V.~Adams, and D.~Paulus, ``{HyKo: A Spectral Dataset
  for Scene Understanding},'' in \emph{Proceedings of the IEEE International
  Conference on Computer Vision Workshops}, 2017, pp. 254--261.

\bibitem{hsiroad}
J.~Lu, H.~Liu, Y.~Yao, S.~Tao, Z.~Tang, and J.~Lu, ``{Hsi Road: A Hyper
  Spectral Image Dataset For Road Segmentation},'' in \emph{2020 IEEE
  International Conference on Multimedia and Expo (ICME)}, 2020, pp. 1--6.

\bibitem{you2019hyperspectral}
S.~You, E.~Huang, S.~Liang, Y.~Zheng, Y.~Li, F.~Wang, S.~Lin, Q.~Shen, X.~Cao,
  D.~Zhang \emph{et~al.}, ``{Hyperspectral city v1. 0 dataset and benchmark},''
  \emph{arXiv preprint arXiv:1907.10270}, 2019.

\bibitem{ipaccess}
\BIBentryALTinterwordspacing
{University of the Basque Country UPV/EHU}, ``{HSI-Drive},'' 2023. [Online].
  Available: \url{https://ipaccess.ehu.eus/HSI-Drive/}
\BIBentrySTDinterwordspacing

\bibitem{gutierrez22}
\BIBentryALTinterwordspacing
J.~Guti\'{e}rrez-Zaballa, K.~Basterretxea, J.~Echanobe, M.~V. Mart\'{\i}nez,
  and I.~del Campo, ``{Exploring Fully Convolutional Networks for the
  Segmentation of Hyperspectral Imaging Applied to Advanced Driver Assistance
  Systems},'' in \emph{Design and Architecture for Signal and Image Processing:
  15th International Workshop, DASIP 2022, Budapest, Hungary, June 20–22,
  2022, Proceedings}.\hskip 1em plus 0.5em minus 0.4em\relax Berlin,
  Heidelberg: Springer-Verlag, 2022, p. 136–148. [Online]. Available:
  \url{https://doi.org/10.1007/978-3-031-12748-9\_11}
\BIBentrySTDinterwordspacing

\bibitem{gutierrez2023chip}
\BIBentryALTinterwordspacing
J.~Gutiérrez-Zaballa, K.~Basterretxea, J.~Echanobe, M.~V. Martínez,
  U.~Martinez-Corral, Óscar Mata-Carballeira, and I.~{del Campo}, ``{On-chip
  hyperspectral image segmentation with fully convolutional networks for scene
  understanding in autonomous driving},'' \emph{Journal of Systems
  Architecture}, vol. 139, p. 102878, 2023. [Online]. Available:
  \url{https://www.sciencedirect.com/science/article/pii/S1383762123000577}
\BIBentrySTDinterwordspacing

\bibitem{gevers2000colour}
T.~Gevers, H.~M. Stokman, and J.~van~de Weijer, ``{Colour Constancy from
  Hyper-Spectral Data},'' in \emph{BMVC}, 2000, pp. 1--10.

\bibitem{finlayson2005removal}
G.~D. Finlayson, S.~D. Hordley, C.~Lu, and M.~S. Drew, ``On the removal of
  shadows from images,'' \emph{IEEE transactions on pattern analysis and
  machine intelligence}, vol.~28, no.~1, pp. 59--68, 2005.

\end{thebibliography}

\newpage
\onecolumn

\end{document}